\documentclass[journal,final,twoside]{IEEEtran}

\ifCLASSOPTIONcompsoc
  \usepackage[nocompress]{cite}
  \usepackage[pdftex]{graphicx}
  \DeclareGraphicsExtensions{.pdf,.jpeg,.png}
  \usepackage[caption=false,font=footnotesize,labelfont=sf,textfont=sf]{subfig}
\else
  \usepackage{cite}
  \usepackage[dvips]{graphicx}
  \DeclareGraphicsExtensions{.eps}
  \usepackage[caption=false,font=footnotesize]{subfig}
\fi
\usepackage{multicol, multirow, booktabs,threeparttable}
\usepackage{amsmath,amsfonts,amssymb,amsthm}
\usepackage{algorithm}
\usepackage{algorithmic}
\usepackage{array}
\usepackage{url}

\usepackage[T1]{fontenc}
\usepackage{xcolor}
\usepackage[short]{optidef}
\usepackage{hhline}
\usepackage{cuted}
\usepackage{stackengine}
\definecolor{r}{rgb}{0.6,0,0}
\definecolor{g}{rgb}{0,0.6,0} 
\usepackage{hyperref}

\newcommand{\vect}[1]{\boldsymbol{#1}}

\newsavebox\mybox

\makeatletter
\newcommand*\dotp{\mathpalette\dotp@{.5}}
\newcommand*\dotp@[2]{\mathbin{\vcenter{\hbox{\scalebox{#2}{$\m@th#1\bullet$}}}}}
\makeatother

\begin{document}
\title{Error-Propagation-Free Learned Video Compression With Dual-Domain Progressive Temporal Alignment}
\author{
Han~Li,
Shaohui~Li,~\IEEEmembership{Member,~IEEE,}
Wenrui~Dai,~\IEEEmembership{Member,~IEEE,}
Chenglin~Li,~\IEEEmembership{Member,~IEEE,}
Xinlong~Pan, Haipeng~Wang,
Junni~Zou,~\IEEEmembership{Member,~IEEE,}
and Hongkai~Xiong,~\IEEEmembership{Fellow,~IEEE}
\thanks{This work was in part supported by National Natural Science Foundation of China, under Grant 62320106003.}
\thanks{Han Li, Chenglin Li, and Hongkai Xiong are with the Department of Electronic Engineering, Shanghai Jiao Tong University, Shanghai 200240, China (e-mail: qingshi9974@sjtu.edu.cn; lcl1985@sjtu.edu.cn; xionghongkai@sjtu.edu.cn).}
\thanks{Wenrui Dai and Junni Zou are with the Department of Computer Science and Engineering, Shanghai Jiao Tong University, Shanghai 200240, China (e-mail: daiwenrui@sjtu.edu.cn; zoujunni@sjtu.edu.cn).}
\thanks{Shaohui Li is with the College of Information Science and Electronic Engineering, Zhejiang University, Hangzhou 310007, China (e-mail: lishaohui.ac@gmail.com).}
\thanks{Xinlong Pan and Haipeng Wang are with the Naval Aviation University, China (e-mail: airadar@126.com; whp5691@126.com).}
}

\markboth{Under Review}%
{Li \MakeLowercase{\textit{et al.}}: Towards Free Error Propagation for Learned Video Compression With Dual-Domain Progressive Temporal Alignment}
\maketitle

\begin{abstract}
Existing frameworks for learned video compression suffer from a dilemma between inaccurate temporal alignment and error propagation for motion estimation and compensation (ME/MC). The separate-transform framework employs distinct transforms for intra-frame and inter-frame compression to yield impressive rate-distortion (R-D) performance but causes evident error propagation, while the unified-transform framework eliminates error propagation via shared transforms but is inferior in ME/MC in shared latent domains. To address this limitation, in this paper, we propose a novel unified-transform framework with dual-domain progressive temporal alignment and quality-conditioned mixture-of-expert (QCMoE) to enable quality-consistent and error-propagation-free streaming for learned video compression. Specifically, we propose dual-domain progressive temporal alignment for ME/MC that leverages coarse pixel-domain alignment and refined latent-domain alignment to significantly enhance temporal context modeling in a coarse-to-fine fashion. The coarse pixel-domain alignment efficiently handles simple motion patterns with optical flow estimated from a single reference frame, while the refined latent-domain alignment develops a Flow-Guided Deformable Transformer (FGDT) over latents from multiple reference frames to achieve long-term motion refinement (LTMR) for complex motion patterns. Furthermore, we design a QCMoE module for continuous bit-rate adaptation that dynamically assigns different experts to adjust quantization steps per pixel based on target quality and content rather than relies on a single quantization step. QCMoE allows continuous and consistent rate control with appealing R-D performance. Experimental results show that the proposed method achieves competitive R-D performance compared with the state-of-the-arts, while successfully eliminating error propagation.
\end{abstract}

\begin{IEEEkeywords}
Learned video compression, unified-transform, progressive alignment, error-propagation, variable-rate coding.
\end{IEEEkeywords}

\IEEEpeerreviewmaketitle

\section{Introduction}
\IEEEPARstart{L}{earned} video compression~\cite{zhao2021enhanced,yang2022advancing,liu2022end,liu2020neural,Lu2019-CVPR,Djelouah2019-ICCV,Feng2020-CVPRW,Hu2021-CVPR,Agustsson2020-CVPR,Lu2020-TPAMI,Li2021-NeurIPS,Lin2020-CVPR,yang2020-CVPR,Guo2021-ARXIV,Sun2021-ARXIV,Sheng2022-TMM,Li2022-ACMMM,Li2023-CVPR,li2024neural,liu2020conditional,shi2022alphavc,mentzer2022vct} has been widely studied along with the rapid development of learned image compression~\cite{Toderici2016-ICLR,Theis2017-ICLR,Balle2018-ICLR,Minnen2018-NeurIPS,Minnen2020-ICIP,Cheng2020-CVPR,wu2021learned,fu2023asymmetric,liu2023learned,li2025on,li2024frequencyaware,li2024image}. Different from the hybrid framework employed by H.264~\cite{Wiegand2003-TCSVT}, H.265~\cite{Sullivan2012-TCSVT}, and VVC~\cite{Bross2021-TCSVT}, learned video compression optimizes all the modules including transform, quantization, entropy coding, and motion estimation and compensation (ME/MC) in an end-to-end fashion to reduce temporal and spatial redundancy in consecutive frames.

Existing methods for learned video compression can be broadly divided into two categories in the sense of neural network based transforms for intra-frame and inter-frame compression, \emph{i.e.}, the separate-transform and unified-transform frameworks. The separate-transform framework such as the DCVC series~\cite{Li2021-NeurIPS,Sheng2022-TMM,Li2022-ACMMM,Li2023-CVPR,li2024neural} employs distinct transform networks (\emph{i.e.}, encoder $g_a$ and decoder $g_s$) for intra-frame (I-frame) and inter-frame (P-frame) compression, and obtains impressive performance that obviously surpasses the hybrid framework. However, the framework usually leverages conditional transforms to produce compact latent representations conditioned on the decoded frames. This causes error propagation due to distorted latent representations with the error accumulated on the decoded frame as condition, and necessitates periodic refreshes~\cite{li2024neural}. Moreover, it causes excessive storage costs using transform networks with significantly different architectures and parameters for intra-frame and inter-frame compression.

Contrary to separate-transform framework, the unified-transform framework avoids error propagation by using a shared transform to produce latent representations for intra-frame and inter-frame compression. However, it cannot sufficiently exploit temporal correlations and is evidently inferior to the separate-transform framework in rate-distortion (R-D) performance. The shared transform could lose fine-grained spatial details in compressed latent representations, and impedes motion estimation/compensation (ME/MC) that is ill-posed without a one-to-one correspondence between frames and requires accurate temporal alignment. For example, VCT~\cite{mentzer2022vct} directly models the probability distribution of current latent conditioned on decoded latents but is degraded in capturing motion patterns without explicit ME/MC. AlphaVC-cI~\cite{shi2022alphavc} employs a simple pixel-to-feature motion predictor for deformable alignment. However, it cannot handle different motion patterns using single-stage temporal alignment.

To address these limitations, we propose a dual-domain progressive temporal alignment framework to reduce temporal redundancy and mitigate the ill-posed ME/MC problem in a coarse-to-fine fashion. Specifically, we disentangle temporal alignment into coarse pixel-domain alignment that offers an initial solution to ME/MC and produces robust motion priors and fine latent alignment that refines the coarse motion priors. The coarse pixel-domain alignment efficiently models simple motion patterns with optical flow estimated from a single reference frame, whereas the fine latent alignment consists of a long-term motion refinement (LTMR) module to refine the optical flow initially estimated in the pixel domain using the priors of multiple reference latents and a flow-guided deformable transformer (FGDT) to obtain the finely aligned latent with the refined optical flow. The coarse-to-fine approach can handle diverse motion patterns and achieve effective temporal context modeling with superior R-D performance.

On such basis, we develop a novel variable-rate learned video compression framework that allows continuous bit-rate adaptation and eliminates error propagation. We integrate the proposed unified-transform compression method with a Quality-Conditioned Mixture-of-Experts (QCMoE) module. By taking the quality embedding as a condition, the QCMoE can flexibly assign different experts to generate quantization steps for each pixel, while different experts focus on different visual characteristics. Compared with existing methods that rely on a single quantization generator to achieve bit-rate adaptation, our QCMoE can achieve more efficient bit-rate adaptation and improved reconstruction quality. With the unified-transform framework, we can adopt a global quality embedding to simultaneously control the quality of reconstructed I- and P-frames for stable reconstruction quality.

The contributions of this paper are summarized below.
\begin{itemize}

\item We propose a dual-domain progressive temporal alignment method for accurate motion estimation and compensation, significantly reducing temporal redundancy of the unified-transform video compression framework.

\item We introduce a quality-conditioned mixture-of-experts (QCMoE) module for continuous bit-rate adaptation, enabling dynamic quantization adjustments based on target quality and content.

\item We develop a learned video compression method under the unified-transform framework, which offers reduced temporal redundancy, consistent and continuous bit-rate adaptation, and eliminated error propagation.

\end{itemize}

Extensive experiments show that the proposed method achieves rate-distortion performance comparable to state-of-the-art approaches. Moreover, it enables consistent and continuous variable-rate coding while effectively preventing error propagation.

The remainder of this paper is organized as follows. Section~\ref{sec:related} briefly reviews the related work. Section~\ref{sec:overview} introduces the proposed unified-transform video coding method that eliminates error propagation and enables consistent and continuous bit-rate adaptation. Section~\ref{sec:progressive} elaborates the proposed dual-domain progressive temporal alignment method that remarkably reduces temporal redundancy. Section~\ref{sec:experiment} presents experimental results. Finally, we draw conclusions in Section~\ref{sec:conclusion}.

\section{Related Work}\label{sec:related}
\subsection{Learned Video Compression}
Benefiting from end-to-end learning, learned video compression has become a widely discussed topic. An early work is DVC~\cite{Lu2019-CVPR} that directly replaces all components of the traditional video coding framework with neural networks. Recent methods can be categorized into two kinds of frameworks, \emph{i.e.}, the separate-transform and unified-transform frameworks. The \textbf{separate-transform framework} employs distinct transform networks (\textit{i.e.}, a pair of encoder and decoder) for I-frame and P-frame compression. For example, DCVC~\cite{Li2021-NeurIPS} adopt the intra-frame transforms for I-frame compression and propose another conditional transforms for P-frame compression to further reduce temporal redundancy. Further improvements on DCVC include temporal context mining~\cite{Sheng2022-TMM}, hybrid spatial-temporal entropy models~\cite{Li2022-ACMMM}, cross-group interaction~\cite{Li2023-CVPR}, and feature modulation~\cite{li2024neural}. The \textbf{unified-transform framework} shares the pair of encoder and decoder for I-frames and P-frames. CECEVC~\cite{liu2020conditional} and VCT~\cite{mentzer2022vct} map all frames in a video sequence to a shared latent space and uses the decoded latent as the prior to model the probability distribution of current frame.  AlphaVC-cI~\cite{shi2022alphavc} transforms the preceding decoded frame  and current frame into a shared latent space, and use a pixel-to-feature motion predictor to achieve temporal alignment. However, the R-D performance of the unified-transform framework is largely inferior to the separate-transform due to insufficient temporal redundancy elimination directly in the shared latent space.

\subsubsection{Error Propagation}
The separate-transform framework suffers inevitable error propagation due to the dependency of inter-frame prediction for reconstruction. Specifically, each P-frame's reconstruction quality relies on its reference frames, any inaccuracies in prior reconstructions directly degrade subsequent predictions. Therefore, current separate-transform methods are weak on long-term prediction. To deal with this problem, Lu~\emph{et~al.}~\cite{lu2020content} propose an error propagation-aware training strategy, which consider the compression performance of multiple consecutive frames instead of a single frame. Based on~\cite{lu2020content}, Ripple~\emph{et~al.}~\cite{Ripple2021-ICCV} further design a strategy that dynamically modulates the reconstruction loss for each frame. However, these methods can only alleviate error propagation rather than eliminate it. In this paper, we adopt the unified transforms for all frames of a video sequence. Since the values of the decoded latent and reconstruction frame is independent on the previous reconstruction frames, our framework is free from the error propagation, as presented in Section~\ref{ssec:overview}.

\subsubsection{Variable-Rate Intra-Frame Compression}
Variable-rate compression is first achieved in learned image compression that can be viewed as intra-frame compression for learned video compression. Toderici~\emph{et~al.}~\cite{Toderici2016-ICLR} use LSTM networks to progressively transmit bits of compressed image. Choi~\emph{et~al.}~\cite{Choi2019-ICCV} adopt a conditional autoencoder and realize variable-rate compression by adjusting the Lagrange multiplier for R-D loss. Similarly, Yang~\emph{et~al.}~\cite{Yang2020-SPL} develop a modulated network to achieve variable-rate image compression. However, these methods require extra complex networks and significantly increase the computational and storage overheads of the whole model. To this end, latent scaling that directly adjusts the quantization step of the latent by scaling factors is considered~\cite{Theis2017-ICLR,Chen2020-ICASSP,Cui2021-CVPR,Balle2021-JSTSP}. Chen~\emph{et~al.}~\cite{Chen2020-ICASSP} use a learnable scalar as the scaling factor to scale the whole latent, while Cui~\emph{et~al.}~\cite{Cui2021-CVPR} scale each channel of latent with independent real values. Content-adaptive quantization strategies have also been explored in recent works. Lee~\emph{et~al.}~\cite{lee2022selective} design a selective method for compressing only significant latent representations in variable-rate image coding. Cai~\emph{et~al.}~\cite{cai2024i2c} incorporate content-adaptive mechanisms to enhance the fidelity of finely-adapted variable-rate image compression.

\subsubsection{Variable-Rate Inter-Frame Compression}
Rippel~\emph{et~al.} \cite{Rippel2019-ICCV} leverage a spatial multiplexer to achieve variable-rate video compression, but it takes  long time to generate the spatial multiplexer, which hinders real-world applications. DCVC-HEM~\cite{Li2022-ACMMM} develops a multi-granularity quantization module. However, this method relies on the hyper-prior entropy model and requires additional bit consumption to transmit element-wise quantization steps. In~\cite{Lin2021-DCC,Lin2021-ICIP,Ripple2021-ICCV}, the idea of latent scaling for intra-frame compression~\cite{Chen2020-ICASSP,Cui2021-CVPR} is transferred to inter-frame compression. Due to the transform functions in intra-frame and inter-frame compression are usually different in these methods, it is hard to design a feasible variable-rate adaptation strategy adapted to both I-frame and P-frame. To deal with it, we formulate a novel variable-rate framework for video compression using our quality-conditioned mixture-of-expert (QC-MoE) module to achieve continuous and consistent bit-rate adaptation for both I-frame and P-frame.

\subsection{Temporal Feature Alignment and Motion Compensation}
Motion compensation in the latent space aims to predict the latent of current frame with that of previously reconstructed frame. It is similar to temporal feature alignment that has been discussed in other video-related tasks~\cite{Zhu2017-CVPR, Xue2019-IJCV, tian2020-CVPR, Deng2020-AAAI, Liang2022-AXIV}, with the key distinction that motion information for video compression needs to be transmitted with minimal bit-budget. For temporal feature alignment, Zhu~\emph{et~al.}~\cite{Zhu2017-CVPR} propagate the deep feature maps of key frame to other frames via a optical flow  to boost the accuracy of video recognition. Xue~\emph{et~al.}~\cite{Xue2019-IJCV} leverage the pre-trained optical flow estimation model SpyNet~\cite{Ranjan2017-CVPR} to generate task-oriented optical flow on various video restoration tasks. Tian~\emph{et~al.}~\cite{tian2020-CVPR} and Deng~\emph{et~al.}~\cite{Deng2020-AAAI} use deformable convolution networks (DCN) to align the feature map of key frame with that of the target frame, thus enhancing the quality of compressed video. Inspired by~\cite{tian2020-CVPR}, Hu~\emph{et~al.}~\cite{Hu2021-CVPR} introduce DCN to implement motion compensation in the latent space for video coding. However, DCN cannot handle scenarios with large and complex motions. Motivated by the success of Transformers in natural language processing (NLP)~\cite{Vaswani2017-NeurIPS,Devlin2018-BERT} and computer vision (CV)~\cite{Dosovitskiy2021-ICLR,Liu2021-ICCV,Qian2022-ICLR,Zhu2022-ICLR,Zou2022-CVPR,Liang2022-AXIV}, we propose a flow-guided deformable Transformer (FGDT) and develop a dual-domain progressive temporal alignment method to reduce the temporal redundancy of consecutive frames.

\section{Proposed Unified-Transform Framework for Learned Video Compression}\label{sec:overview}
In this section, we elaborate the proposed unified-transform framework that achieves consistent and continuous variable-rate coding and eliminates error propagation. Table~\ref{tab:symbols} summarizes the frequently used notations.

\subsection{Overview}\label{ssec:overview}
Let $\{\vect{X}_0,\cdots,\vect{X}_t,\cdots,\vect{X}_{n-1}\}$ denote a group of pictures (GOP) in the video sequence to be compressed, where the first frame $\vect{X}_0$ is the only I-frame and the remaining are P-frames. \figurename~\ref{fig:overview} illustrates the intra-frame and inter-frame coding for the I-frames and P-frames, respectively. 

\begin{table}[!t]
\renewcommand{\baselinestretch}{1.0}
\renewcommand{\arraystretch}{1.0}
\setlength{\tabcolsep}{3pt}
\setlength{\abovecaptionskip}{0pt}
\centering
\caption{Summary of frequently used notations and their descriptions.}\label{tab:symbols}
\centering
\begin{tabular}{l|c|l}
\hline
& Notation & Description\\
\hline
\multirow{6}{*}{\rotatebox{90}{\parbox{1.1cm}{\centering Basic Notation}}}
& $g_\mathrm{a}(\cdot)$ & Encoder with learnable parameters $\theta_\mathrm{a}$ \\
& $g_\mathrm{s}(\cdot)$ & Decoder with learnable parameters $\theta_\mathrm{s}$\\
& $\vect{X}_t$ & Source (the $t$-th frame of the input GOP )\\
& $\vect{Y}_t$ & The $t$-th latent, $\vect{Y}_t = \operatorname{QCMoE}(g_\mathrm{a}(\vect{X}_t))$ \\

& $\widehat{\vect{Y}}_{t}$ & The $t$-th reconstructed latent \\

& $\widehat{\vect{Y}}_t$ & The $t$-th reconstructed source, $\widehat{\vect{X}}_t = g_\mathrm{s}(\operatorname{i-QCMoE}(\widehat{\vect{Y}}_t))$ \\

\hline
\multirow{4}{*}{\rotatebox{90}{\parbox{1.5cm}{\centering QCMoE}}}
& $\vect{Y}^{'}$ & unscaled latent\\
& $\vect{Y}$ & scaled latent\\
& $\mathcal{Q}$ & the set of quality embedding\\
& $\vect{q}$ & quality embedding\\
\hline
\multirow{4}{*}{\rotatebox{90}{\parbox{1.9cm}{\centering Progressive Alignment}}}

&$\vect{V}_{t-1\rightarrow t}$& Optical flow between  $\vect{X}_t$ and$\vect{\widehat{X}}_{t-1}$\\ 
&$\vect{\widehat{V}}_{t-1\rightarrow t}$& Reconstructed optical flow between  $\vect{X}_t$ and $\vect{\widehat{X}}_{t-1}$\\ 
&$\vect{\widehat{O}}_{t-1\rightarrow t}$& Downsampled  optical flow \\
&$\vect{o}_{t-1\rightarrow t}$& Predicted refined offset for downsampled optical flow \\
&$\vect{\overline{O}}_{t-1\rightarrow t}$& Refined downsampled optical flow \\
&$\vect{\widetilde{X}_t}$& Compensated frame of $\widehat{\vect{X}}_{t-1}$ warped by $\vect{\widehat{V}}_{t-1\rightarrow t}$ \\
&$\vect{\widetilde{Y}}_t$& The $t$-th coarsely aligned latent \\
&$\vect{\check{Y}}_t$& The $t$-th finely aligned latent \\
\hline 
\end{tabular}
\end{table}

\textbf{Intra-Frame Compression.} We encode the I-frame with an intra-frame coding scheme. To enable variable-rate adaptation, we develop Quality-Conditioned Mixture-of-Experts (QCMoE) with a quality embedding $\vect{q}$ for dynamic latent scaling. As shown in \figurename~\ref{fig:overview}(b), we first use an encoder $g_\mathrm{a}$ and a QCMoE to transform $\vect{X}_0$ into the latent $\vect{Y}_0$ given the quality embedding $\vect{q}$. $\vect{Y}_{0}$ is then sent to the intra-frame entropy model and encoded into the bitstream via arithmetic coding. At the decoder side, the reconstructed latent $\widehat{\vect{Y}}_{0}$ is transformed into the reconstructed I-frame $\widehat{\vect{X}}_{0}$ using inverse QCMoE (\textit{i}-QCMoE) and decoder $g_\mathrm{s}$. The QCMoE controls latent scaling with the quality embedding $\vect{q}$ to enable continuous rate-distortion trade-off with a single model.

\textbf{Inter-Frame Compression.} The P-frames are encoded via an inter-frame coding scheme. As shown in \figurename~\ref{fig:overview}(a), we leverage the previously reconstructed frame $\widehat{\vect{X}}_{t-1}$ and previously reconstructed latents $\widehat{\vect{Y}}_{t-1}, \widehat{\vect{Y}}_{t-2}, \widehat{\vect{Y}}_{t-3}$ from decoded buffer to implement progressive alignment, achieving effective temporal redundancy reduction for encoding current frame $\widehat{\vect{X}}_t$.
Specifically, we obtain the alignment of current latent in a coarse-to-fine manner, where $\vect{\widetilde{Y}}_t$ and $\vect{\check{Y}}_t$ are the coarsely aligned and finely aligned latents, respectively. The prediction $\vect{\widetilde{Y}}_t$ and $\vect{\check{Y}}_t$ are served as temporal prior in the inter-frame entropy model to encode the $\vect{Y}_t$. At the decoder side, the reconstructed latent $\widehat{\vect{Y}}_{t}$ is transformed back into the reconstructed frame $\widehat{\vect{X}}_t$ using the inverse QCMoE (\textit{i}-QCMoE) and the decoder $g_\mathrm{s}$.

Note that the transform networks (\emph{i.e.}, encoders, decoders, QCMoE, and \textit{i}-QCMoE) are shared across intra-frame and inter-frame compression such that I-frames and P-frames are transformed into the same latent space. In addition to reducing model parameters, this unified transform framework can consistently control the reconstruction quality and bit-rates of all the frames in a video sequence with a single quality embedding $\vect{q}$. Moreover, we design a multi-stage training schedule to adequately train each module of the proposed framework. In the remaining part of this section, we elaborate our variable-rate coding method using QCMoE. Besides, the proposed uniform-transform framework is free from error-propagation, as detailed below. Let the a GOP of the original video sequence be $\{\vect{X}_0, \vect{X}_1, \dots, \vect{X}_{n-1}\}$, where $\vect{X}_0$ is the I-frame and the rest are P-frames. In our unified-transform framework, a shared encoder $g_a$ and decoder $g_s$ are used for all frames. For the $t$-th frame, the latent representation is $\vect{Y}_t = g_a(\vect{X}_t)$ and the reconstructed frame is $\widehat{\vect{X}}_t = g_s(\widehat{\vect{Y}}_t)$, where $\widehat{\vect{Y}}_t$ is the quantized latent representation,  $i.e., \widehat{\vect{Y}}_t= \left\lfloor \vect{Y}_t\right\rceil$. While inter-frame entropy coding model the distribution of $\widehat{\vect{Y}}_t$ conditioned on the previous decoded frame $\widehat{\vect{X}}_{t-1}$ and decoded latents $\widehat{\vect{Y}}_{t-1}, \widehat{\vect{Y}}_{t-2},$ and $\widehat{\vect{Y}}_{t-3}$, the value of quantized latent $\widehat{\vect{Y}}_t$ depends solely by the current frame's latent $\vect{Y}_t$ and quantization operation $\lfloor \cdot \rceil $,  not on previous reconstructions. In this way, $\widehat{\vect{X}}_t$ is computed exclusively from $\widehat{\vect{Y}}_t$ through $g_s(\cdot)$, with no functional dependence on previous reconstructions. Thus, the quantization error $ \left\lfloor \vect{Y}_t\right\rceil -\vect{Y}_t$ affects only the current frame and will not propagate to subsequent frames.

\begin{figure}[!t]
\renewcommand{\baselinestretch}{1.0}
\setlength{\abovecaptionskip}{0pt}
\centering
\includegraphics[width=3.5 in]{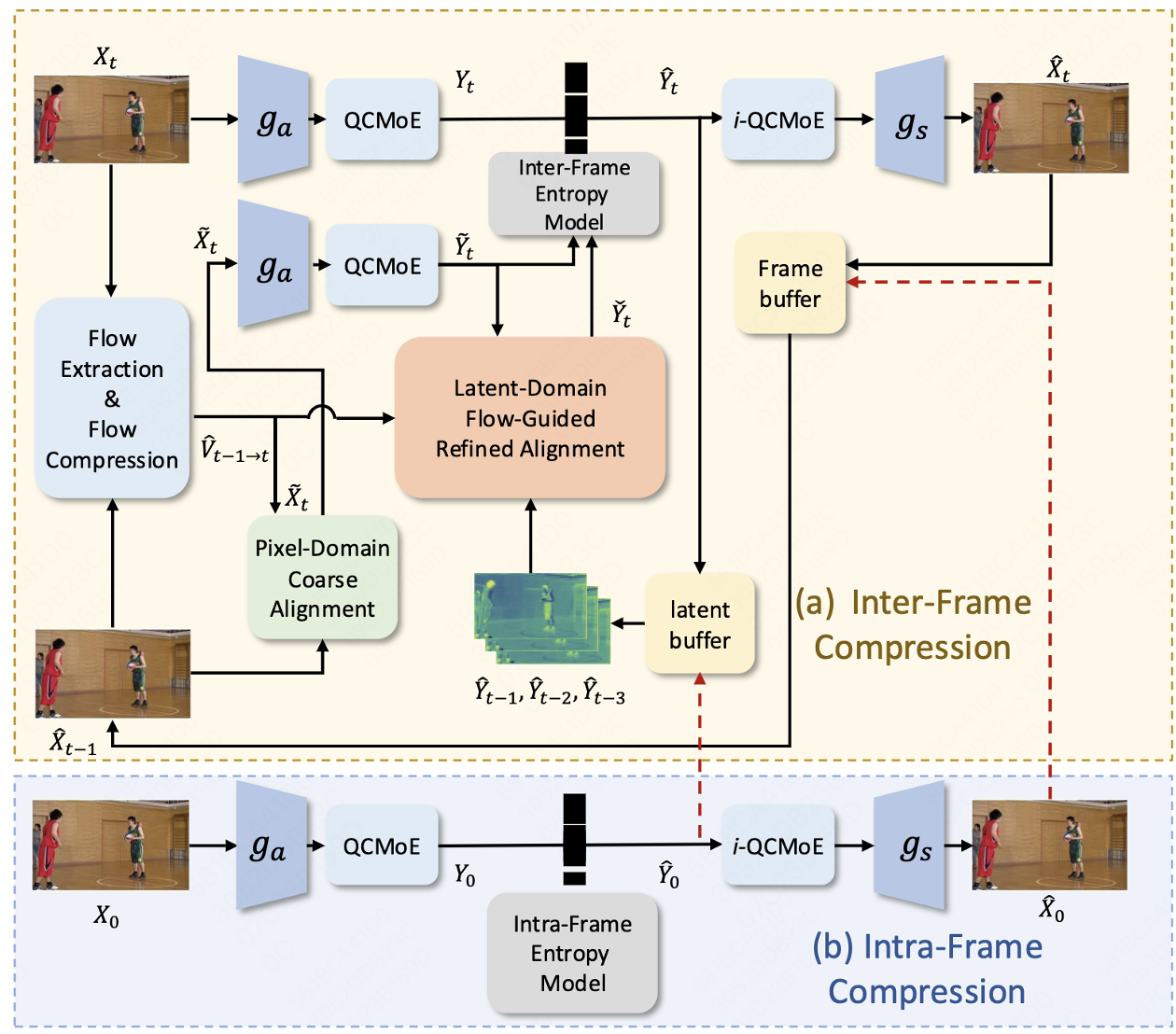}
\caption{Overview of the proposed unified-transform video compression framework, which leverages the dual-domain progressive temporal alignment for improved R-D performance and supports ability for continuous variable-rate coding.}
\label{fig:overview}
\end{figure}

\subsection{Quality-Conditioned Mixture-of-Experts for Latent Scaling}\label{sec:latent-scaling}
Learned image and video coding methods achieve variable-rate compression by adjusting quantization steps via latent scaling. Naive latent scaling is first validated for learned image compression in~\cite{Chen2020-ICASSP} to achieve variable-rate compression using a single nonlinear transform. Channel-wise latent scaling~\cite{Cui2021-CVPR} extends the limited bit-rate range produced by naive latent scaling. Content-aware latent scaling is developed in the DCVC series~\cite{Li2022-ACMMM,Li2023-CVPR,li2024neural} to enhance rate-distortion performance, but requires bit-rate overhead for decoding quantization parameters and relies on a fixed module to generate the quantization step. 
In this section, we first analyze the principles of naive and channel-wise latent scaling operations and then elaborate the proposed latent scaling method that improves the R-D performance via a light quality-conditioned mixture-of-experts module. For simplicity, we omit the time subscript $t$ to unify the symbols for I-frames and P-frames.

\subsubsection{Naive Latent Scaling}
Let $\vect{Y}^{\prime}=g_a(\vect{X})$ be the $t$-th latent in a GOP and $\mathbb{S}=\{s|s_{\mathrm{min}}\leq s\leq s_{\mathrm{max}}\}$ be a set of available scaling factors. The quantization process combined with naive latent scaling (using $s\in \mathbb{S}$) can be formulated in three steps.

\begin{itemize}
\item[\textit{a}.] Scaling: $\vect{Y} = \vect{Y}^{\prime}\cdot s$;
\item[\textit{b}.] Quantization: $\widehat{\vect{Y}} = \left\lfloor \vect{Y}\right\rceil$;
\item[\textit{c}.] Re-scaling: $\widehat{\vect{Y}}^{\prime} = \widehat{\vect{Y}}\cdot s^{-1}$.
\end{itemize}
Here, the notation $\left\lfloor\cdot\right\rceil$ denotes the rounding operation that returns the integer closest to the input. $\vect{Y}$, $\widehat{\vect{Y}}$, and $\widehat{\vect{Y}}^{\prime}$ represent the scaled latent, the discrete latent, and the rescaled quantized latent, respectively. Adjusting the scaling factor controls the quantization error between $\vect{Y}^{\prime}$ and $\widehat{\vect{Y}}^{\prime}$, which is equivalent to changing the quantization step of uniform quantization. Suppose we denote the uniform quantization with a quantization step $\Delta$ as $\widehat{\vect{Y}}^{\prime} = \Delta\cdot\left\lfloor\vect{Y}^{\prime}/\Delta\right\rceil$. Then the quantization process equipped with naive latent scaling is identical to the uniform quantization with a quantization step $\Delta = 1/s$. When the value of scaling factor increases, the quantization error between $\widehat{\vect{Y}}^{\prime}$ and $\vect{Y}^{\prime}$ becomes smaller and the reconstruction error for current frame decreases. Besides, the continuous scaling factor set $\mathbb{S}$ is commonly available even though the compression model are trained with only several discrete scaling factors.

\subsubsection{Channel-Wise Latent Scaling}
The $C$ channels of $\vect{Y}^{\prime}$ are unbalanced in the influence on reconstruction distortion. Thus, the scalar $s$ in naive latent scaling can be replaced by a vector $\vect{m}\in \mathbb{R}^C$ to allow a unique quantization step for each channel. Suppose that there is a set of scaling vectors $\mathcal{M}=\left\{\vect{m}_1,\vect{m}_2,\cdots,\vect{m}_N\right\}$ where each scaling vector corresponds to a target rate. The quantization process incorporated with channel-wise latent scaling is as follows.
\begin{itemize}
\item[\textit{a}.] Scaling: $\vect{Y} = \vect{Y}^{\prime}\odot \vect{m}$;
\item[\textit{b}.] Quantization: $\widehat{\vect{Y}} = \left\lfloor \vect{Y}\right\rceil$;
\item[\textit{c}.] Re-scaling: $\widehat{\vect{Y}}^{\prime} = \widehat{\vect{Y}}\odot\vect{m}^{-1}$.
\end{itemize}
Here, $\vect{m}\in\mathcal{M}$ is the selected scaling vector, $\odot$ represents the channel-wise multiplication, and $\vect{m}^{-1}$ is a vector where each element is the inverse of that in $\vect{m}$.

\subsubsection{Proposed QCMoE}
\label{Sec:our-AOM}
Previous methods used a fixed module to generate quantization steps, which cannot effectively identify the different scene and bitrate requirements. In contrast, the proposed QCMoE dynamically allocate different submodules to adjust quantization steps based on diverse visual characterization and quality condition, which can  preserve critical details while reducing bitrate more effectively.

\begin{figure}[!t]
\renewcommand{\baselinestretch}{1.0}
\setlength{\abovecaptionskip}{0pt}
\centering
\includegraphics[width=0.9\linewidth]{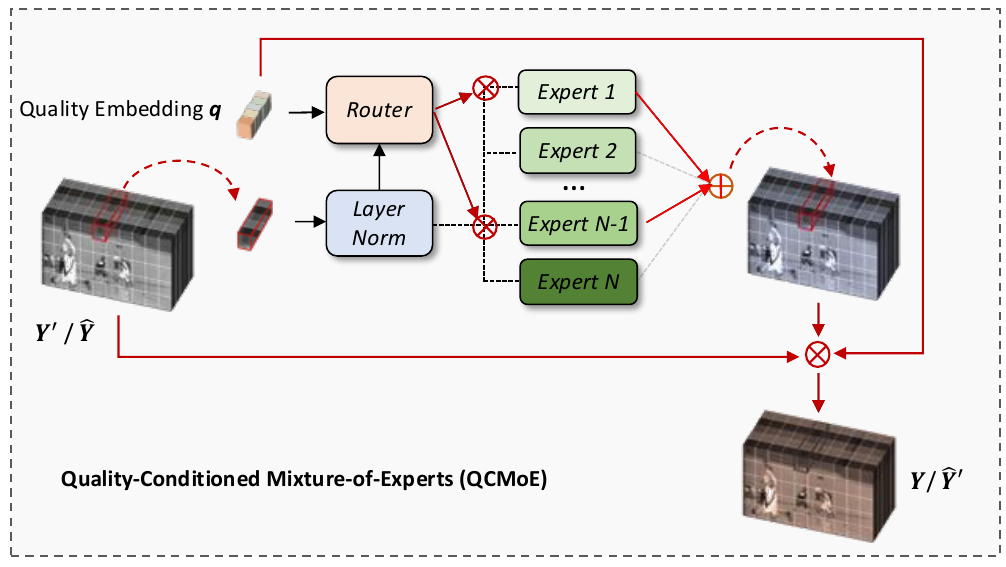}
\caption{The architecture of QCMoE. Each pixel of $\vect{Y}^{\prime}$ is assigned to different scaling expert conditioned on the quality embedding to generate the corresponding scaling factor, \emph{i.e.}, quantization step. Then $\vect{Y}^{\prime}$ is scaled by the scaling factor to obtain the scaled latent $\vect{Y}$.
}
\label{fig:QCMOE}
\end{figure}

As depicted in \figurename~\ref{fig:QCMOE}, QCMoE consists of a group of $M$ experts $f_1, \cdots, f_M$ along with a router $\mathcal{R}$. Each expert is a multilayer perceptron (MLP) designed to handle distinct visual characteristics and bitrate requirements. The router $\mathcal{R}$ is an MLP that allocates specialized experts to generate quantization steps conditioned on the latent and quality embedding.

Specifically, we employ a top-$K$ gating router to assign the experts. As presented in \eqref{eq:qcmoe}, the output of QCMoE is a weighted sum of the top $K$ experts from $M$ expert candidates, where the weights are determined by the router.
\begin{equation}\label{eq:qcmoe}
\begin{aligned}
\operatorname{QCMoE}(\vect{Y}^{\prime},\vect{q}) & = \vect{q}·\sum_{k=1}^{K} \mathcal{R}(\vect{Y}^{\prime},\vect{q})_k \cdot f_k(\vect{Y}^{\prime}), \\
\mathcal{R}(\vect{Y}^{\prime},\vect{q}) &= \operatorname{TopK}(\operatorname{softmax}(\mathcal{G}(\vect{Y}^{\prime}+\vect{q}), K)), \\
\operatorname{TopK}(v, K) &= 
\begin{cases} 
v & \text{if } v \text{ is in the top } K \text{ elements} \\
0 & \text{otherwise}
\end{cases},
\end{aligned}
\end{equation}
where $\mathcal{G}$ represents the MLP layer for the router. The $\operatorname{softmax}(·)$ together with $\operatorname{TopK}(~\cdot~, K)$ sets all elements of the vector to zero except the elements with the largest $K$ values.
\begin{align}
\textit{a}.~ &\text{Scaling:} ~\vect{Y} = \vect{Y}^{\prime}\otimes \operatorname{QCMoE}(\vect{Y}^{\prime},\vect{q}),\label{eq:cals-1}\\
\textit{b}.~ &\text{Quantization:} ~\widehat{\vect{Y}} = \left\lfloor \vect{Y}\right\rceil,\label{eq:cals-2}\\
\textit{c}.~ &\text{Re-scaling:} ~\widehat{\vect{Y}}^{\prime} = \widehat{\vect{Y}}\otimes [i\operatorname{-QCMoE}(\widehat{\vect{Y}},\vect{q})]^{-1}.\label{eq:cals-3}
\end{align}
Here, $\otimes$ denotes the element-wise multiplication and $[i\operatorname{-QCMoE}(\widehat{\vect{Y}},\vect{q})]^{-1}$ is a matrix whose elements are the inverse of those in $i\operatorname{-QCMoE}(\widehat{\vect{Y}},\vect{q})$. Note that $\operatorname{QCMoE}$ and \textit{i}$\operatorname{-QCMoE}$ share identical structures but have distinguishing parameters. To balance performance and complexity, we set $M$ to 6 and $K$ to 2. The quality embedding $\vect{q}$ is chosen from the set $\mathcal{Q} =\{ \vect{q}_1,\cdots,\vect{q}_N\}$ based on the given quality index for training, and is interpolated to achieve continuous bitrate adjustment during inference.

\begin{figure}[!t]
\renewcommand{\baselinestretch}{1.0}
\setlength{\abovecaptionskip}{0pt}
\centering
\includegraphics[width=3.5 in]{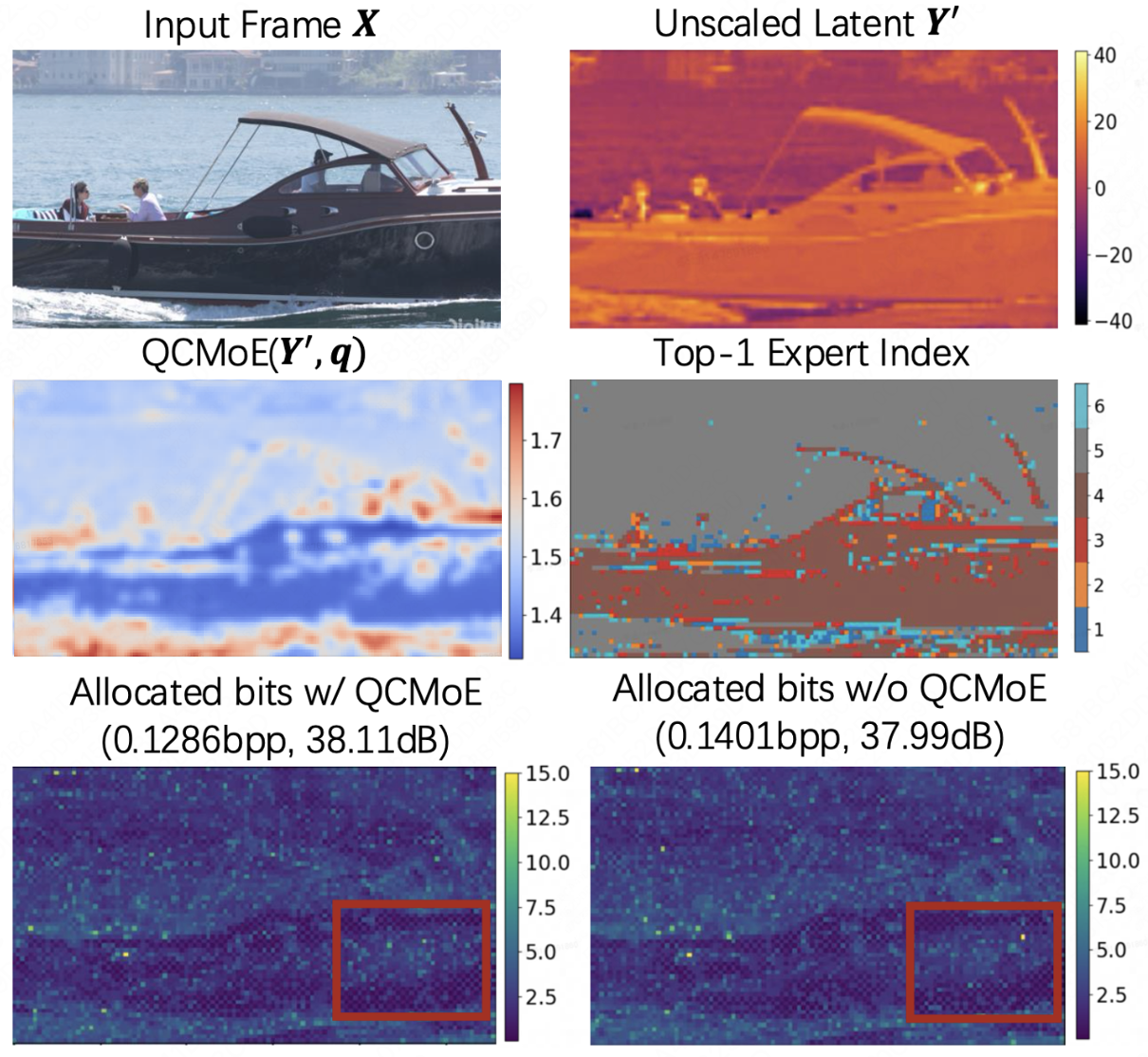}
\caption{Visual examples of the proposed QCMoE. We use the channel with maximal bit budget for illustration. The \textbf{left of middle row} shows the scaling factor map generated by QCMoE at encoder side, \emph{i.e.}, $\operatorname{QCMoE}(\vect{Y}^{\prime},\vect{q})$. The \textbf{right of middle row} shows the index map of Top-1 expert for each pixel. Compared with getting rid of the QCMoE module, our methods can achieve 8\% bit-rate saving, while bring 0.12\,dB improvement for PSNR. }\label{fig:vis_aom}
\end{figure}

\figurename~\ref{fig:vis_aom} visualizes an example of the proposed QCMoE for the channel with maximal bit budget. From the visualization of scaling factor map  $\operatorname{QCMoE}(\vect{Y}^{\prime},\vect{q})$, we find that the texture regions are scaled with greater factors while the smooth regions are with smaller factors. This fact means that the texture regions are quantized with smaller quantization steps and smooth regions with larger quantization steps. As a result, the overall quantization errors can be balanced by considering the contents in the texture and smooth regions. We also present the index of assigned top-1 expert for each pixel. We find that the 4-th and 5-th experts focus more on the smoothed region while other experts focus on the edge of the entity. The proposed QCMoE is a plug-and-play and lightweight module that can be seamlessly embedded to most existing fixed-rate learned image and video compression models to enable continuous variable-rate compression.

\section{Dual-Domain Progressive Temporal Alignment for Inter-Frame Compression}
\label{sec:progressive}
Based on the proposed unified-transform framework with QCMoE, we present a novel progressive dual-domain alignment framework that hierarchically reduces temporal redundancy through coarse-to-fine motion modeling. The dual-domain progressive temporal alignment can precisely resolve complex motion characteristics like non-rigid deformation and large-displacement scene transition. Specifically, we first extract optical flow between current frame and the previously decoded frame. Then, we perform efficient coarse alignment in the pixel domain to capture simple motion patterns (\emph{e.g.}, rigid translations), establishing initial correspondences between frames. Further, we employ refined flow-guided alignment in the latent domain which consists of a long-term motion refinement (LTMR) to refine the optical flow estimated in the pixel domain and a flow-guided deformable transformer (FGDT) to obtain the finely aligned latent.

\subsection{Flow Extraction and Compression}
Following DCVC-HEM~\cite{Li2022-ACMMM}, for $t>1$, we employ lightweight SpyNet \cite{Ranjan2017-CVPR} to estimate optical flow 
$\vect{V}_{t-1\rightarrow t}$ between current frame  $\vect{X}_t$ and its preceding reconstructed frame $\vect{\widehat{X}}_{t-1}$ as the estimated motion vector in the pixel domain.
$\vect{V}_{t-1\rightarrow t}$ is losslessly compressed using an autoencoder incorporated with a hyperprior entropy model and obtain the reconstructed optical flow $\widehat{\vect{V}}_{t-1\rightarrow t}$. 

\subsection{Coarse Pixel-domain Alignment}
The coarse pixel-domain alignment module establishes initial motion correspondences through efficient pixel-domain warping operations. As shown in \figurename~\ref{fig:overview}, the previously decoded frame $\vect{\widehat{X}}_{t-1}$ is warped by the optical flow $\widehat{\vect{V}}_{t-1\rightarrow t}$ to generate the compensated frame $\vect{\widetilde{X}}_{t}$.
\begin{equation}
\widetilde{\vect{X}}_{t} = \operatorname{Warp}(\vect{\widehat{X}}_{t-1},\widehat{\vect{V}}_{t-1\rightarrow t}).
\end{equation}
$\vect{\widetilde{X}}_{t}$ is then encoded via our unified encoder $g_a$ and QCMoE to obtain $\vect{\widetilde{Y}_{t}}$ that serves as a coarsely aligned latent effectively capturing dominant motion patterns. Subsequently, we develop latent-domain refinement for $\vect{\widetilde{Y}_{t}}$ to address the misalignment due to flow estimation inaccuracies and compression artifacts.

\begin{figure}[!t]
\renewcommand{\baselinestretch}{1.0}
\setlength{\abovecaptionskip}{0pt}
\centering
\includegraphics[width=0.9\linewidth]{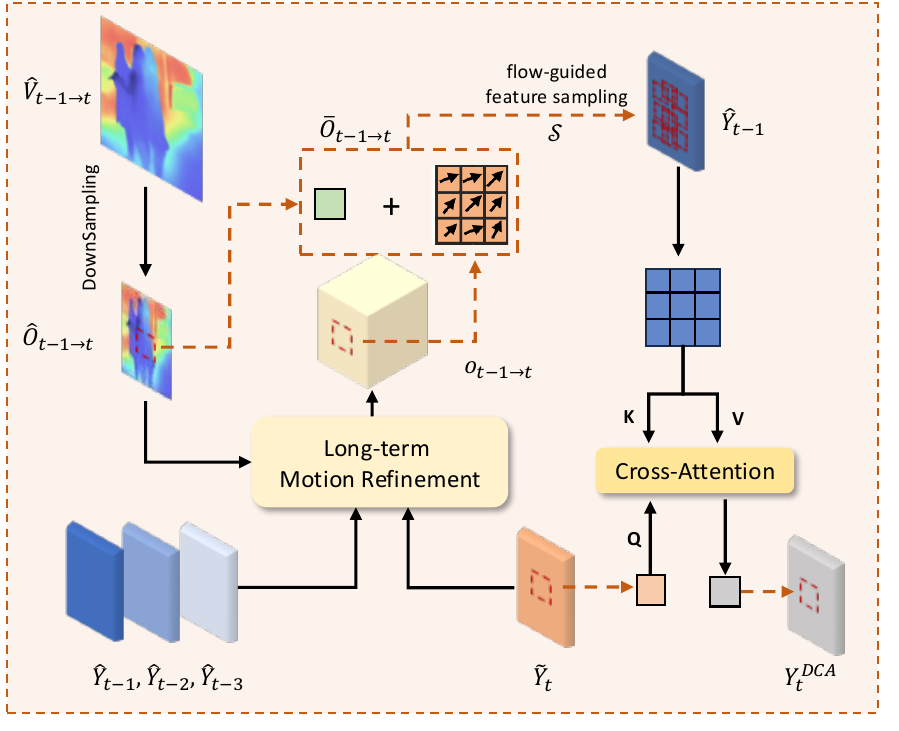}
\caption{The architecture of flow-guided deformable cross-attention (FG-DCA).}\label{fig:fg-dca}
\end{figure}

\subsection{Flow-Guided Refined Latent-domain Alignment}
We develop a flow-guided refined latent-domain alignment to refine the coarsely aligned latent $\vect{\widetilde{Y}}_{t}$ using multiple previously decoded latents from the latent buffer.  

\subsubsection{Long-Term Motion Refinement}
Directly utilizing optical flow estimated in the pixel domain for alignment in the latent domain is suboptimal, as the motion characteristics in the latent domain differ significantly from those in the pixel domain. To address this discrepancy, we introduce the LTMR module for refining the motion in the latent domain.
Specifically, we first downsample the optical flow to match the size of latent, \emph{i.e.}, $\widehat{\vect{O}}_{t-1 \rightarrow t} = \operatorname{downsample}(\widehat{\vect{V}}_{t-1 \rightarrow t})$, and then predict $L$ refined offset from the concatenation of coarsely aligned latent $\vect{\widetilde{Y}}_{t}$, several previously decoded latents $\widehat{\vect{Y}}_{t-1},\widehat{\vect{Y}}_{t-2},\widehat{\vect{Y}}_{t-3}$, and the downsampled optical flow $\widehat{\vect{O}}_{t-1 \rightarrow t}$ along channel dimension using a lightweight group convolution layer ($\operatorname{GroupConv}$) for efficient long-term temporal modeling. The refined offset is obtained by:
\begin{equation}
\vect{o}_{t\!-\!1\! \rightarrow t}\!=\! \operatorname{GroupConv}(
\operatorname{Concat}(\widetilde{\vect{Y}}_{\!t},\widehat{\vect{Y}}_{\!t\!-\!1},\widehat{\vect{Y}}_{\!t\!-\!2},\widehat{\vect{Y}}_{\!t\!-\!3},\widehat{\vect{O}}_{\!t\!-\!1\!\rightarrow t})).
\end{equation}
where $\vect{o}_{t\!-\!1\! \rightarrow t}^{i}\!\in \mathbb{R}^{L\times 2}$ is the $L$ refined offsets for the $i$-th position of latent. Then we sum up the refined offsets and the downsampled optical flow to obtain the refined refined optical flow $\overline{\vect{O}}_{t-1 \rightarrow t}$, which is used as the predicted location for the subsequent deformable transformer.
\begin{equation}
    \overline{\vect{O}}_{t-1 \rightarrow t} = \widehat{\vect{O}}_{t-1 \rightarrow t} +\vect{o}_{t-1 \rightarrow t},
\end{equation}
where $\vect{\overline{O}}_{t\!-\!1\! \rightarrow t}^{i}\!\in \mathbb{R}^{L\times 2}$ is the $L$ predicted locations for the $i$-th position of latent.

\subsubsection{Flow-Guided Deformable Transformer}
For accurate alignment towards current latent $\widehat{\vect{Y}}_t$, we employ a flow-guided deformable cross-attention (FG-DCA) as depicted in \figurename~\ref{fig:fg-dca}. Specifically, we sample the relevant features from $\widehat{\vect{Y}}_{t-1}$ according to the refined optical flow (\emph{i.e.,} predicted locations) $\overline{\vect{O}}_{t-1 \rightarrow t}$. For simplicity, we define the queries $\vect{Q}$, keys $\vect{K}$ and values $\vect{V}$ as follows:
\begin{align}
    \vect{Q} &= \widetilde{\vect{Y}}_{t}\vect{P}^Q,  \\
    \vect{K} &= \mathcal{S}(\widehat{\vect{Y}}_{t-1}\vect{P}^K, \overline{\vect{O}}_{t-1 \rightarrow t}), \\
  \vect{V} &=\mathcal{S}(\widehat{\vect{Y}}_{t-1}\vect{P}^V, \overline{\vect{O}}_{t-1 \rightarrow t}),
\end{align}
where $\mathcal{S}$ denotes the flow-guided feature sampling operation~\cite{zhu2017flow}, and we use bilinear interpolation for sampling to make it differentiable. $\vect{Q}^i\in \mathbb{R}^{1\times C}$ is the  projected feature from the $i$-th position of  coarsely aligned latent $\widetilde{\vect{Y}}_{t}$. $\vect{K}^i\in \mathbb{R}^{L\times C}$ and $\vect{V}^i\in \mathbb{R}^{L\times C}$ are the corresponding projected features that are sampled from $L$ locations of previously decoded latent $\widehat{\vect{Y}}_{t-1}$. $\vect{P}^Q \in \mathbb{R}^{C\times C}$,  $\vect{P}^K \in \mathbb{R}^{C\times C}$ and $\vect{P}^V \in \mathbb{R}^{C\times C}$ are the project matrices for query, key, and value, respectively.
The output of DCA $\vect{Y}_{t}^{\mathrm{DCA}}$ is 
\begin{equation}\label{eq:MHCA}
\vect{Y}_{t}^{\mathrm{DCA}} \!=\! \operatorname{cross-attn}(\vect{Q},\vect{K},\vect{V})\!=\!\operatorname{softmax}\!\left(\frac{\vect{Q}\vect{K^T}}{\sqrt{C}}\right)\!\vect{V}.
\end{equation}

Subsequently, a multi-layer perceptron (MLP) is used for feature transformation and the final output is fused with the coarsely aligned latent $\widetilde{\vect{Y}}_{t-1}$ to obtain the finely aligned latent $\check{\vect{Y}}_t$. The whole process is formulated as below.
\begin{align}
    \vect{Y}_t^{\prime} &= \operatorname{FG-DCA}(\vect{\widetilde{Y}}_{t},\widehat{\vect{Y}}_{t-1},\overline{\vect{O}}_{t-1\rightarrow t}) +\widehat{\vect{Y}}_{t-1}\label{eq:FG-DCA}, \\
    \vect{Y}_t^{\prime\prime} & = \operatorname{MLP}(\vect{Y}_t^{\prime} )+\vect{Y}_t^{\prime}, \\
    \check{\vect{Y}}_t& = \operatorname{Conv}(\operatorname{Concat}(\vect{Y}_t^{\prime\prime},\vect{\widetilde{Y}}_{t})).
\end{align}

\subsection{Inter-Frame Entropy Model}
We model each element $\widehat{Y}_{t,i}$ of the quantized latent $\widehat{\vect{Y}}_t$ with a single Gaussian distribution with mean $\mu_i$ and scale $\sigma_i$. To improve efficiency and accuracy of probability prediction, we use the spatial-channel context model as our inter-frame entropy model, which includes the spatial-channel-condition model and the hyper-prior model. As a complementary of side information (hyper-prior) and spatial-channel prior, we further extract temporal prior information from the coarsely aligned latent $\widetilde{\vect{Y}}_t$ and finely aligned latent $\check{\vect{Y}}_t$ using a temporal context module that consists of two 3$\times$3 group convolutional layers and a non-linear unit. In this way, the predicted Gaussian parameters are functions of learned parameters of hyper-decoder, spatial-channel-conditional module, and temporal context module (denoted by $\vect{\theta}_{\mathrm{hd}},\vect{\theta}_{\mathrm{sccm}}$, and $\vect{\theta}_{\mathrm{tcm}}$).

\begin{table*}[!t]
\setlength{\abovecaptionskip}{0pt}
\centering
\caption{Detailed information of our multi-stage training schedule. ``BS'' denotes batch size and ``LR'' denotes the initial learning rate. Please refer to Section~\ref{Sec:multi-stage-traning-schedule} for detailed explanation.}\label{tab:multi-stage-training}
\begin{tabular}{@{}lc|ccccc@{}}
\toprule
Stages & Components &Frames&  Loss function& BS & LR &\# of steps\\ 
\midrule
I: Intra-Frame Training & Intra-Frame Compression Scheme & 1& Rate-Distortion: Eq.~\eqref{eq:rdloss—intra}&16& 1\emph{e}-4 & 0.5M\\ \hline
\multirow{3}{*}{II: Inter-Frame Training}  & Coarse Alignment & 2&Alignment MSE: Eq.~\eqref{eq:coarse_alignment}&16 & 1\emph{e}-4 & 0.5M \\
 & Refined Alignment & 4&Alignment MSE: Eq.~\eqref{eq:refine_alignment}&16 & 1\emph{e}-4 & 0.5M \\ 
& Inter-Frame Entropy Model & 4&Rate: Eq.~\eqref{eq:rdloss—inter}&16 & 1\emph{e}-4 & 0.5M \\\hline
\multirow{3}{*}{III: Joint Training}  & All &4& Rate-Distortion: Eq.~\eqref{eq:rdloss—joint}&8&2\emph{e}-5 & 2.5M \\ 
 & All &4& Rate-Distortion: Eq.~\eqref{eq:rdloss—joint}&8&1\emph{e}-5 & 0.5M \\ 
& All &4& Rate-Distortion: Eq.~\eqref{eq:rdloss—joint}&8&1\emph{e}-6 & 0.2M \\ 
\bottomrule
\end{tabular}
\vspace{-18pt}
\end{table*}
\begin{figure*}[!t]
\renewcommand{\baselinestretch}{1.0}
\setlength{\abovecaptionskip}{3pt}
\centering
\subfloat[PSNR on UVG]{\includegraphics[width=2.1 in]{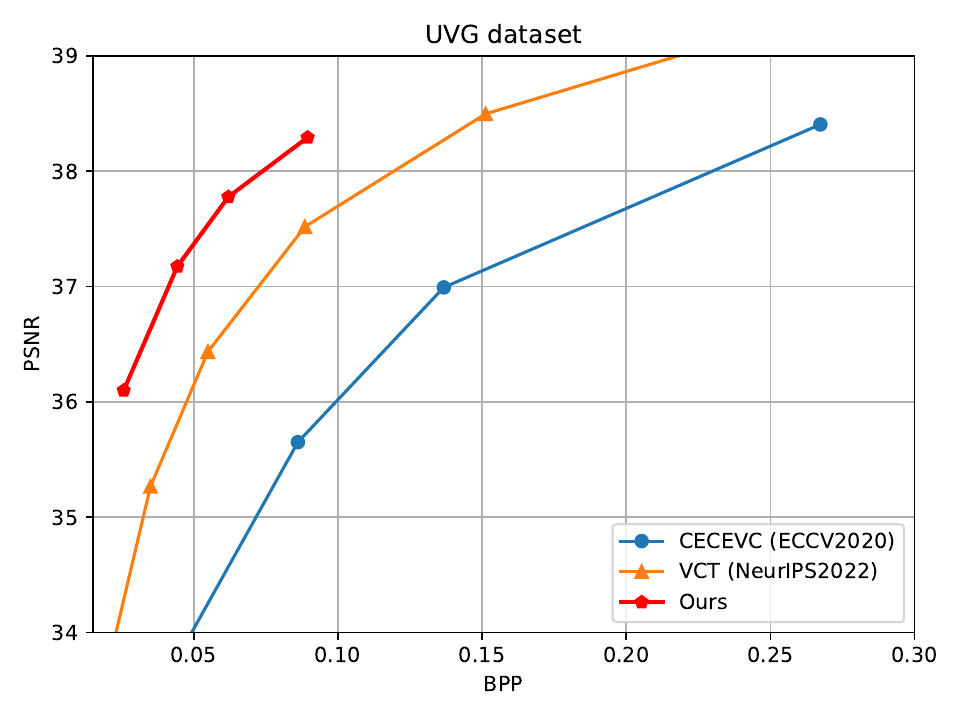}}   
\subfloat[PSNR on MCL-JCV]{\includegraphics[width=2.1 in]{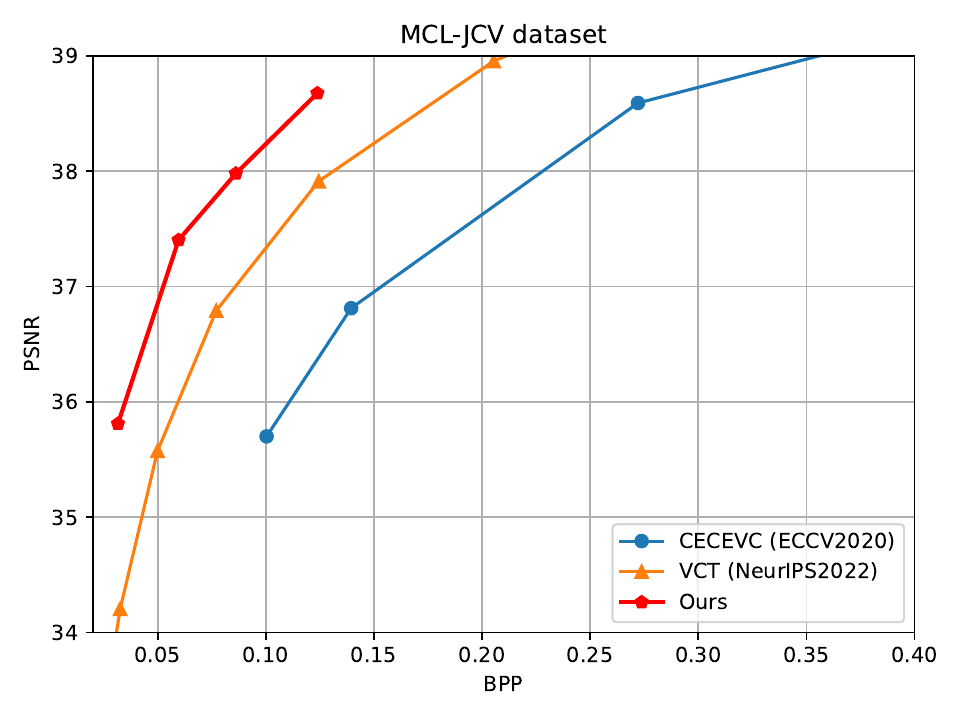}}
\subfloat[PSNR on HEVC Class B]{\includegraphics[width=2.1 in]{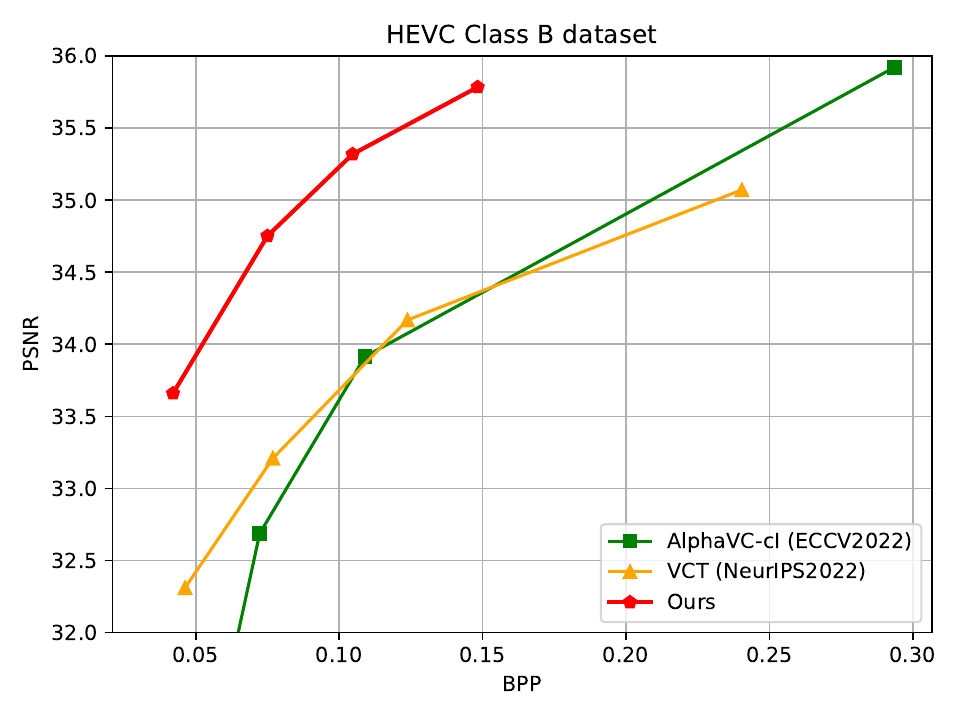}}\
\caption{Comparison with SOTA methods of unified-transform framework on the UVG, MCL-JCV, and HEVC Class B datasets under the PSNR metric.}\label{fig:inter_results_with_unified}
\end{figure*}

\section{Experiments}
\label{sec:experiment}
\subsection{Datasets and Implementation Details}\label{Sec:multi-stage-traning-schedule}
\subsubsection{Training Dataset}
We train both the intra-frame and inter-frame compression schemes on the Vimeo-90k dataset~\cite{Xue2019-IJCV} that contains 89,800 video clips with each clip having 7 frames of 448$\times$256 pixels. We randomly crop the video sequences to the resolution of 256$\times$256 for training.
\subsubsection{Test Datasets}
The proposed method is evaluated on several benchmark datasets, \emph{i.e.}, UVG~\cite{Mercat2020-UVG}, MCL-JCV~\cite{Wang2016-MCL}, and HEVC test sequences (Class B, C, D, and E)~\cite{Sullivan2012-TCSVT}. The UVG dataset contains 7 high frame rate videos with the resolution of 1920$\times$1080. The MCL-JCV dataset is widely used for video quality evaluation and consists of 30 video sequences with the resolution of 1920$\times$1080. The HEVC dataset contains 16 videos with diverse resolutions from 416$\times$240 to 1920$\times$1080.

\subsubsection{Multi-stage Training Schedule}
We adopt a multi-stage training schedule to optimize each module of our framework step by step for stable training. All the training stages are implemented using Adam optimizer~\cite{Kingma2015-ICLR} on a single NVIDIA 4090 GPU. Table~\ref{tab:multi-stage-training} summarizes the multi-stage training schedule along with the loss function for each stage and Table~\ref{tab:multi-stage-training} provides the detailed hyper-parameters, as elaborated below.

\textbf{Stage I: Intra-Frame Training.} We first train the variable-rate intra-frame compression scheme (\emph{i.e.}, ``Intra-Frame'' in Table~\ref{tab:multi-stage-training}) shown in \figurename~\ref{fig:overview}(a) by optimizing the encoder, decoder, QCMoE, \textit{i}-QCMoE, intra-frame entropy model, and the set of quality embedding. Specifically, we adopt the nonlinear transforms and entropy model of \textit{ELIC}~\cite{he2022elic} and load the pre-trained weights \footnote{\url{https://github.com/VincentChandelier/ELiC-ReImplemetation/tree/main}}  for faster convergence.

To achieve variable-rate adaptation, we follow the strategy in~\cite{Cui2021-CVPR} to implement R-D optimization using multiple Lagrange multipliers $\lambda_i$ with the index $i=1,\cdots,4$. The distortion $D$ is measured by mean square error (MSE) or multi-scale structural similarity (MS-SSIM), and  $\lambda_i$ is selected from $\Lambda_\mathrm{MSE}=\{0.020,0.036,0.070,0.130\}$ for minimizing MSE and $\Lambda_\mathrm{MS-SSIM}=\{26.24,45.00,86.10,155.60\}$ for minimizing MS-SSIM. The objective of Stage I training is 
\begin{equation}\label{eq:rdloss—intra}
\min\,R+\lambda_i D,
\end{equation}
where $R$ is the bit-rate for encoding the input frame. During training, $i$ is randomly selected from 1 to 4 in each step to determine the quality embedding $\vect{q}_i\in \mathcal{Q}$ and the Lagrange multiplier $\lambda_i\in\Lambda_\mathrm{MSE}/\Lambda_\mathrm{MS-SSIM}$. $\vect{q}_i$ is optimized with corresponding $\lambda_i$ to adapt to different bit-rates.

\begin{figure*}[!t]
\renewcommand{\baselinestretch}{1.0}
\setlength{\abovecaptionskip}{3pt}
\centering
\subfloat[PSNR on UVG]{\includegraphics[width=2.1 in]{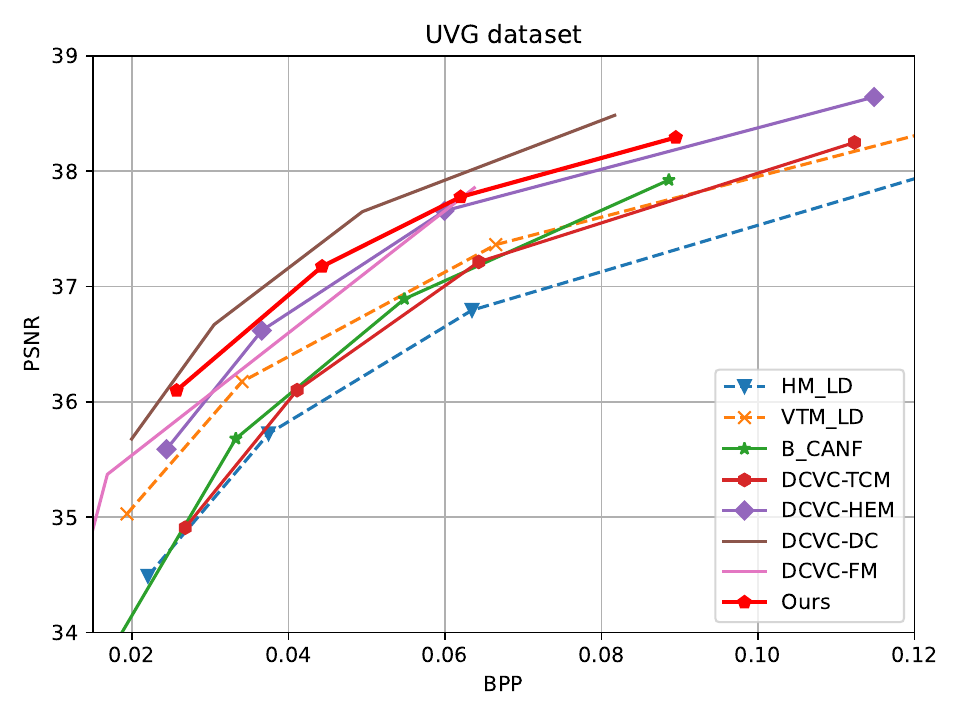}}   
\subfloat[PSNR on MCL-JCV]{\includegraphics[width=2.1 in]{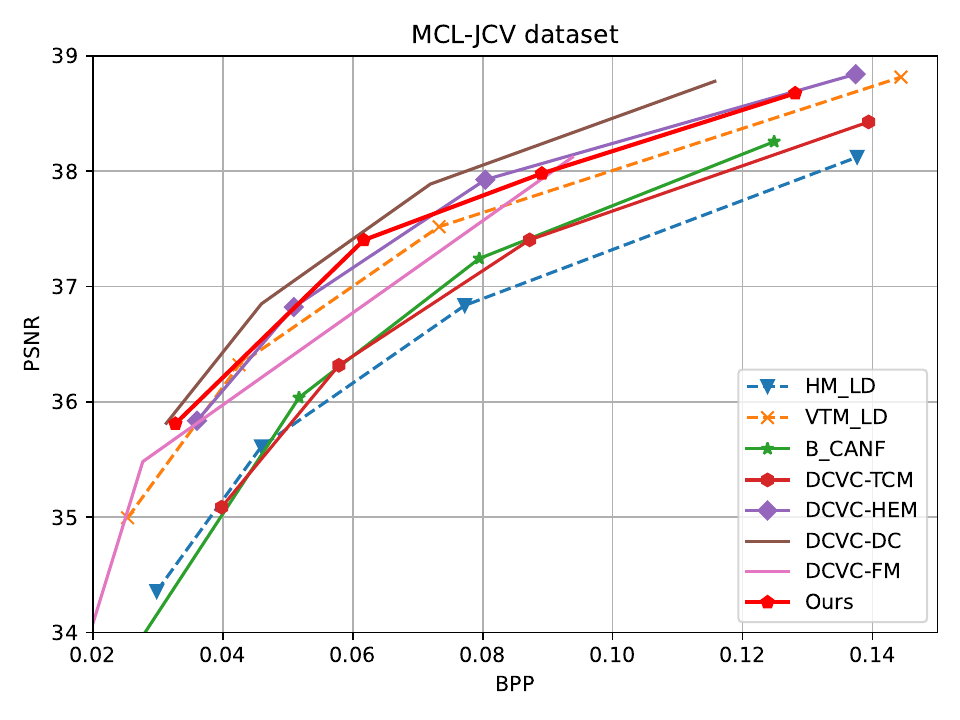}}
\subfloat[PSNR on HEVC Class B]{\includegraphics[width=2.1 in]{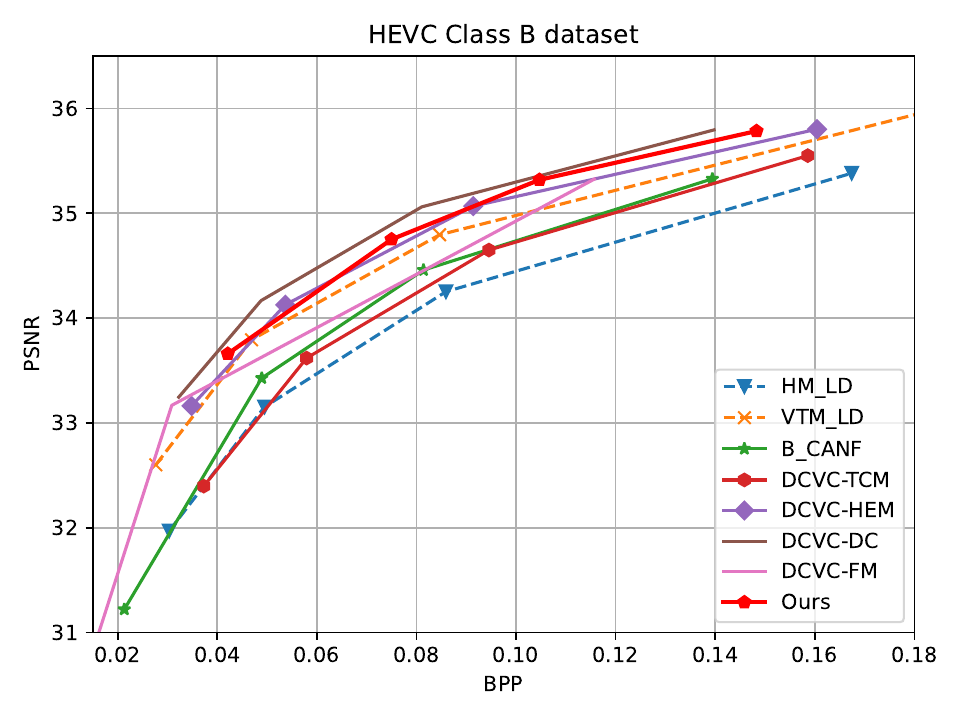}}\\
\vspace{-12pt}
\subfloat[PSNR on HEVC Class C]{\includegraphics[width=2.1 in]{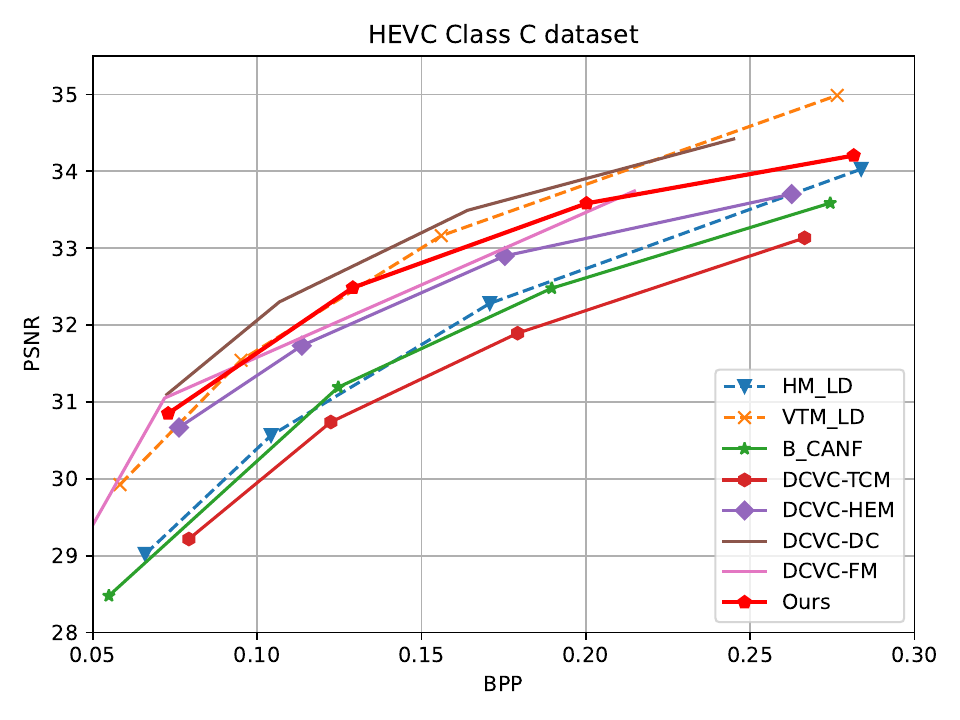}}
\subfloat[PSNR on HEVC Class D]{\includegraphics[width=2.1 in]{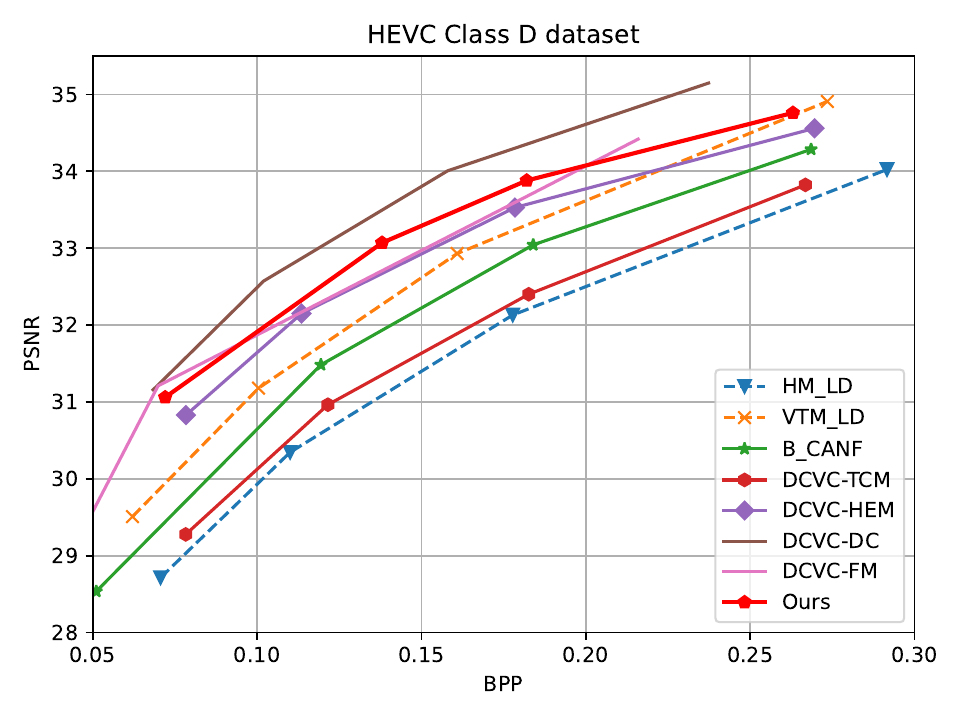}}
\subfloat[PSNR on HEVC Class E]{\includegraphics[width=2.1 in]{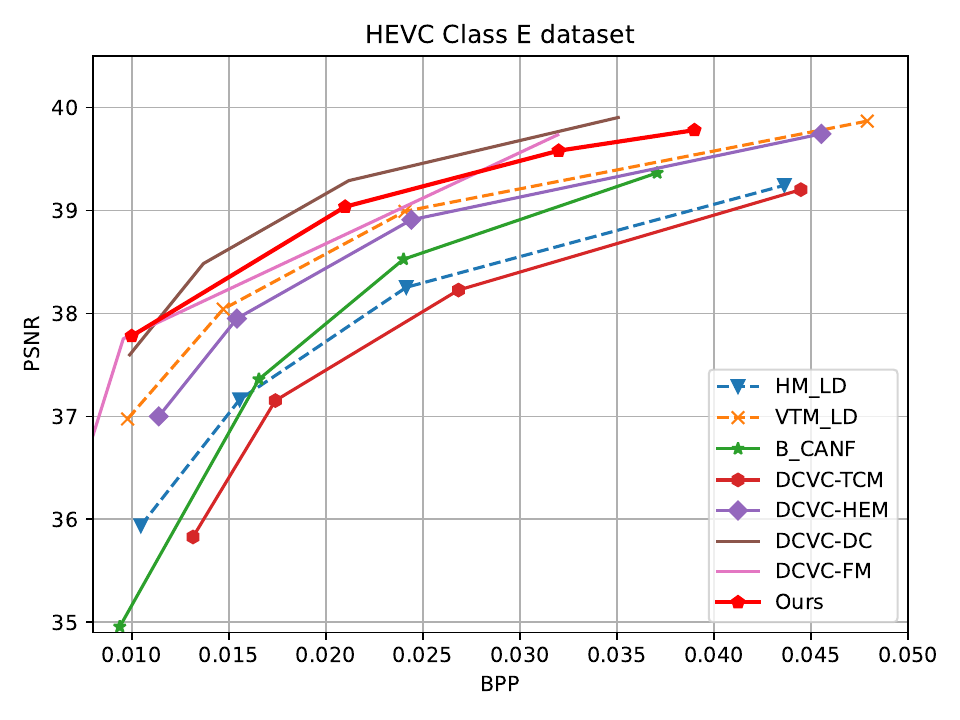}}
\caption{Comparison with SOTA learning-based methods of separate-transform framework and traditional video codecs on the UVG, MCL-JCV and HEVC Class B, C, D, E datasets. PSNR is used for evaluations.}\label{fig:inter_results_psnr}
\vspace{-9pt}
\renewcommand{\baselinestretch}{1.0}
\centering
\subfloat[MS-SSIM on UVG]{\includegraphics[width=2.1 in]{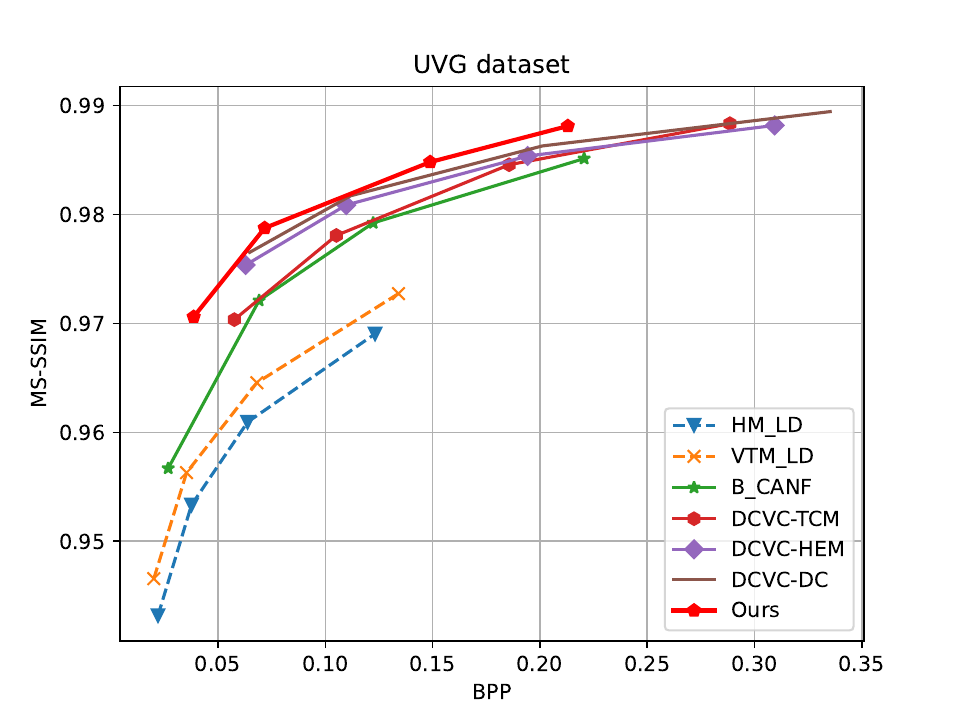}}   
\subfloat[MS-SSIM  on MCL-JCV]{\includegraphics[width=2.1 in]{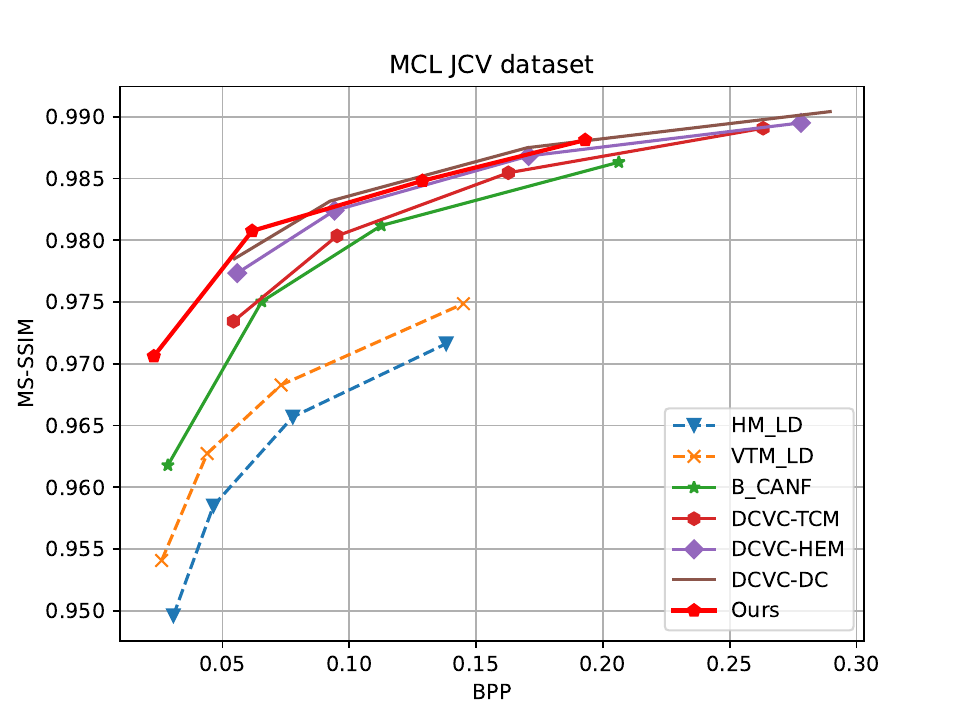}}
\subfloat[MS-SSIM  on HEVC Class B]{\includegraphics[width=2.1 in]{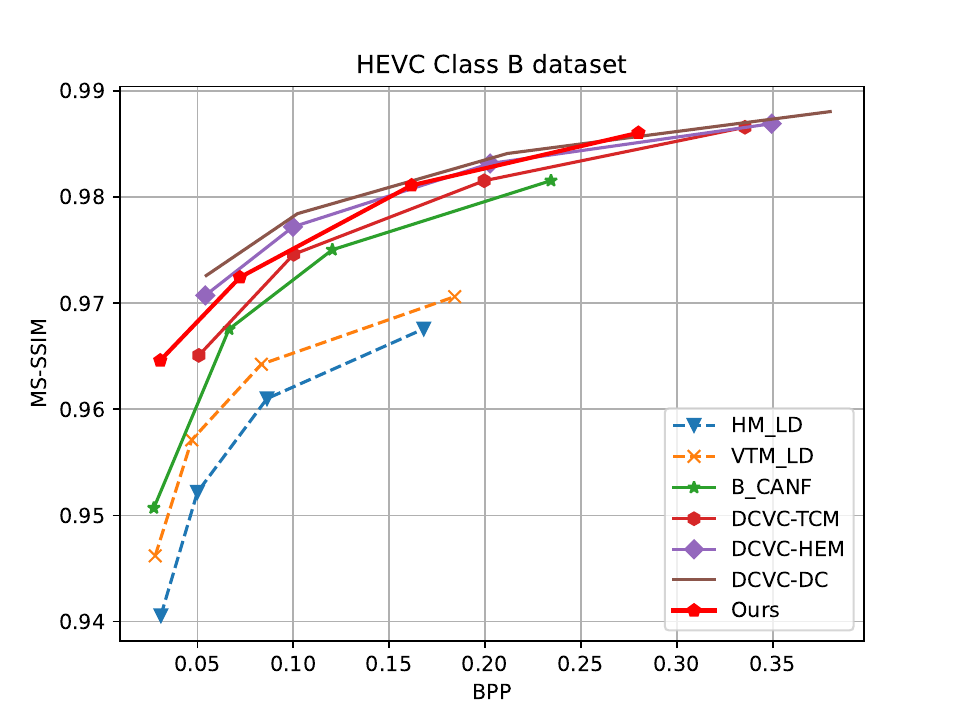}}\\
\vspace{-12pt}
\subfloat[MS-SSIM on HEVC Class C]{\includegraphics[width=2.1 in]{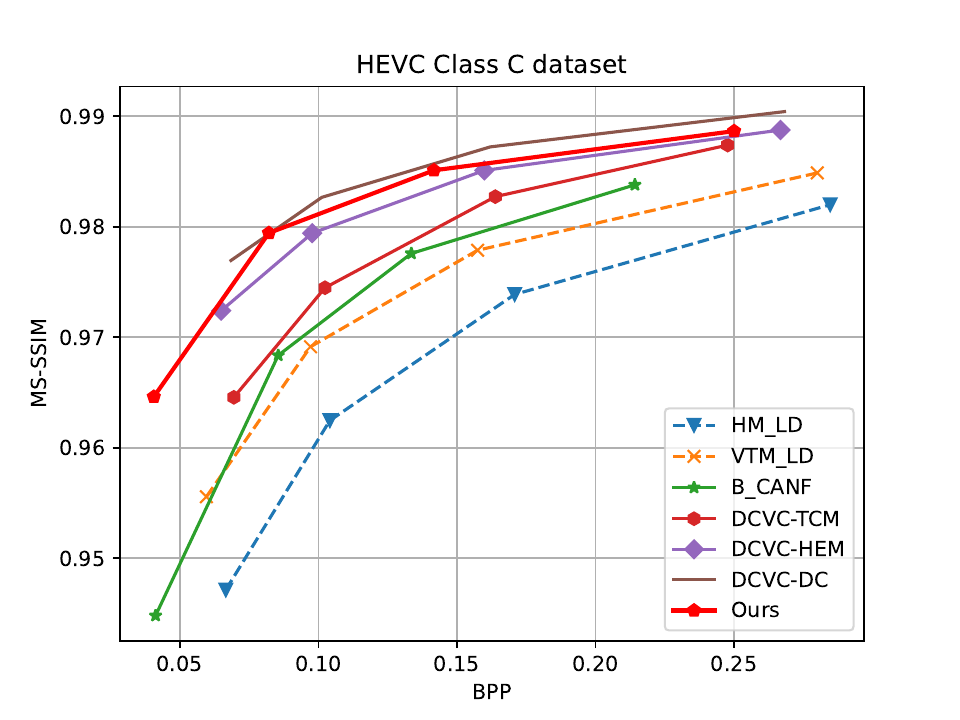}}
\subfloat[MS-SSIM on HEVC Class D]{\includegraphics[width=2.1 in]{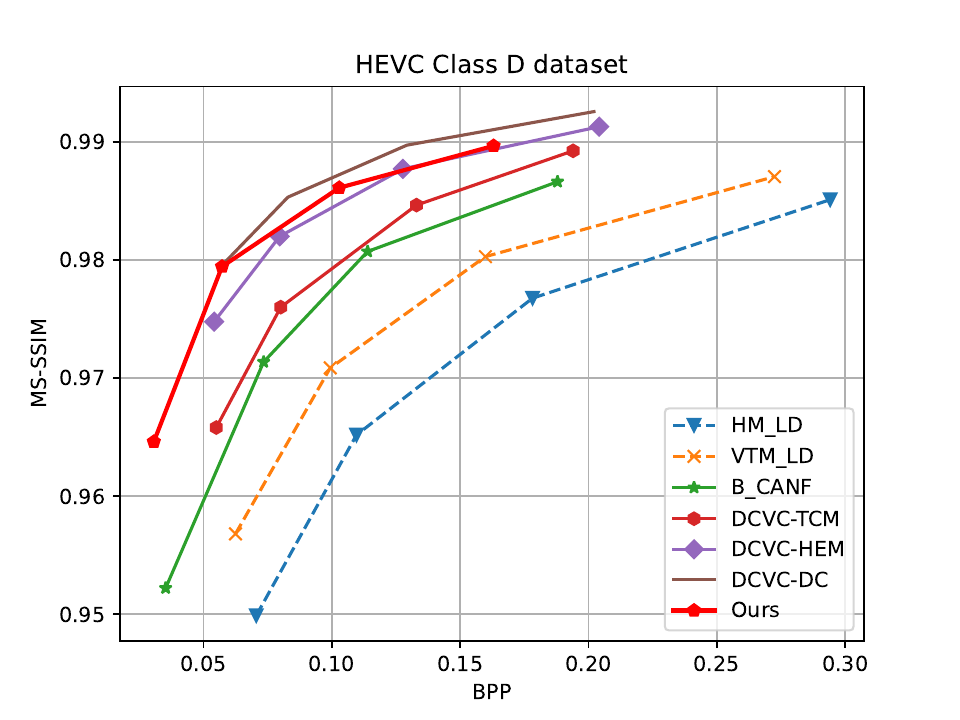}}
\subfloat[MS-SSIM on HEVC Class E]{\includegraphics[width=2.1 in]{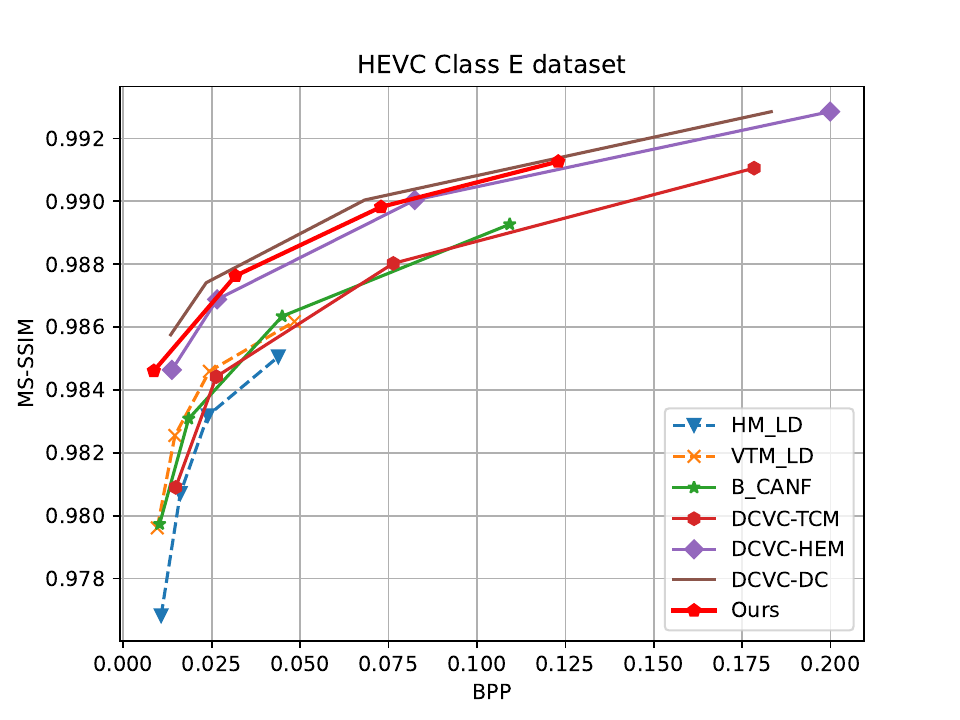}}
\caption{Comparison with SOTA learning-based methods of separate-transform framework and traditional video codecs on the UVG, MCL-JCV and HEVC Class B, C, D, E datasets. MS-SSIM is used for evaluations.}\label{fig:inter_results_ssim}
\end{figure*}

\textbf{Stage II: Inter-Frame Training.} The inter-frame compression model is trained by freezing the parameters of intra-frame compression model optimized in Stage I. We progressively train the submodules of coarse pixel-domain alignment, flow-guided refined latent-domain alignment, and inter-frame entropy model for inter-frame compression. For flow extraction and compression, we adopt the same architectures as DCVC-HEM~\cite{Li2022-ACMMM} initialized with the official pretrained weights for SpyNet and motion encoder and decoder. For coarse and refined alignment, we set the number of refined offsets $L$ to 32 and employ an alignment loss to optimize the MSE between the aligned latents $\vect{\widetilde{Y}}_t/\vect{\check{Y}}_t$ and current latent $\vect{Y}_t$.
\begin{align}
&\text{Coarse Alignment:}~~\min\,\mathrm{MSE}\,(\widetilde{\vect{Y}}_{t},\vect{Y}_{t}),\label{eq:coarse_alignment} \\
&\text{Refined Alignment:}~~\min\,\mathrm{MSE}\,(\check{\vect{Y}}_{t},\vect{Y}_{t}), \label{eq:refine_alignment}
\end{align}
We then optimize the inter-frame entropy model by considering only the bit-rates $R_P$ of P-frames, since the reconstruction quality of a P-frame is independent of prior reconstructions and has been determined during intra-frame training via the optimized unified transform, as presented in Section~\ref{ssec:overview}.
\begin{equation}\label{eq:rdloss—inter}
\min\,R_P.
\end{equation}
Here, $R_P$ includes the bit-rates of latent, motion information, and hyperprior. Similar to the training process in Stage I, we randomly select the quality index from 1 to 4 to adapt inter-frame compression to diverse reconstruction qualities.

\textbf{Stage III: Joint Training.} We jointly train all the parameters of the whole model in an end-to-end fashion. The model is optimized with a R-D loss using $\lambda_i\in\Lambda_\mathrm{MSE}/\Lambda_\mathrm{MS-SSIM}$.
\begin{equation}\label{eq:rdloss—joint}
\min\,R_I+R_P+\lambda_i (D_I +D_P),
\end{equation}
where $R_I$ and $R_P$ denote the bit-rates of the I-frame and P-frames, and $D_I$ and $D_P$ corresponding distortion, respectively.

\begin{table*}[!t]
\renewcommand{\baselinestretch}{1.0}   
\renewcommand{\arraystretch}{1.0}
\setlength{\tabcolsep}{12pt}
\setlength{\abovecaptionskip}{0pt}
\centering
\begin{threeparttable}
\caption{BD-rate (\%)  measured with PSNR. The anchor is VTM. 96 frames with GOP size equal to 32.}\label{tab:inter_results_psnr}
\begin{tabular}{@{}l|ccccccc@{}}
\toprule
 & UVG & MCL-JVC & HEVC Class B & HEVC Class C & HEVC Class D & HEVC Class E &Average \\ 
\midrule
VTM~\cite{Bross2021-TCSVT} &0  &0  &0  &0  & 0 & 0 & 0\\
HM~\cite{Sullivan2012-TCSVT} &34.2  &43.8 & 41.8 & 44.0 & 37.3 &  49.6  &38.6\\
B-CANF~\cite{chen2023b}  &13.9  &30.6 & 21.8  & 47.9 & 10.8 &34.9 &   24.5 \\
DCVC-TCM~\cite{Sheng2022-TMM} &16.5 &32.6 & 29.2 & 67.7 & 30.2 & 66.5  & 39.4\\
DCVC-HEM~\cite{Li2022-ACMMM} & -17.7 &-5.7& -5.2 & 17.3 & -8.6 &  8.6 &10.3 \\
DCVC-DC~\cite{Li2023-CVPR} & -33.2  & -15.5& -15.8 & -7.5 & -28.4 & -25.2  & -20.6\\
DCVC-FM~\cite{li2024neural} & -26.8 & -7.2& -8.3 &  -5.6& -26.3 & -24.7 & -16.3 \\
Proposed &  -24.4 & -10.4& -7.7 &4.9& -16.6 &  -16.4 &-11.7\\ 
\bottomrule
\end{tabular}
\end{threeparttable}
\end{table*}

%
\begin{table*}[!t]
\renewcommand{\baselinestretch}{1.0}   
\renewcommand{\arraystretch}{1.0}
\setlength{\tabcolsep}{12pt}
\setlength{\abovecaptionskip}{0pt}
\centering
\begin{threeparttable}
\caption{BD-rate (\%)  measured with MS-SSIM. The anchor is VTM. 96 frames with GOP size equal to 32.}\label{tab:inter_results_psnr_ms-ssim}
\begin{tabular}{@{}l|ccccccc@{}}
\toprule
 & UVG & MCL-JVC & HEVC Class B & HEVC Class C & HEVC Class D & HEVC Class E &Average \\ 
\midrule
VTM~\cite{Bross2021-TCSVT} &0  &0  &0  &0  & 0 & 0 & 0\\
HM~\cite{Sullivan2012-TCSVT} &27.5  &38.9 & 38.7 & 38.2 & 35.0 & 45.7&25.5\\
B-CANF~\cite{chen2023b}  &-37.8  &-43.4 & -31.9  & -9.3 & -27.7 &7.6 &   -20.3 \\
DCVC-TCM~\cite{Sheng2022-TMM} &-47.7 &-57.4 & -53.2 & -20.7& -36.6 & 14.9  & -28.1\\
DCVC-HEM~\cite{Li2022-ACMMM} & -62.9 &-70.2& -71.0 & -43.8& -55.2 &  -50.0 &-50.4 \\
DCVC-DC~\cite{Li2023-CVPR} &-65.9 & -74.3& -76.9 & -53.9 & -63.6 & -67.1  &  -66.9\\
Proposed & -67.5 & -65.7 & -64.5& -50.1 & -58.5 & -63.2 & -64.0 \\ 
\bottomrule
\end{tabular}
\begin{tablenotes}
\item[$\ast$] The MS-SSIM optimized weights of DCVC-FM are not open-sourced.
\end{tablenotes}
\end{threeparttable}
\end{table*}

\subsection{Performance Evaluation}
\subsubsection{R-D Performance}
We validate the efficacy of the proposed method by comparing it with traditional codecs (\emph{i.e.}, HEVC~\cite{Sullivan2012-TCSVT} and VVC~\cite {Bross2021-TCSVT}), and recent learning-based methods using separate transform (\emph{i.e.}, B-CANF~\cite{chen2023b},   DCVC~\cite{Li2021-NeurIPS}, DCVC-TCM~\cite{Sheng2022-TMM}, DCVC-HEM~\cite{Li2022-ACMMM}, DCVC-DC~\cite{Li2023-CVPR}, and DCVC-FM~\cite{li2024neural}) and unified transform (\emph{i.e.} CECEVC~\cite{liu2020conditional}, AlphaVC-cI~\cite{shi2022alphavc}, and VCT~\cite{mentzer2022vct}) for intra-frame and inter-frame compression. We adopt HM-16.20 using the \emph{encoder\_lowdelay\_main.cfg} configuration with forward prediction for HM for HEVC, and VTM-13.2 using the \emph{encoder\_lowdelay\_vtm.cfg} configuration for VVC. For all the methods, distortion is measured on the decoded sequences within the RGB444 color space. The proposed method determines the quality factor $q$ corresponding to the target bit-rate for both I-frame and P-frame in the same video sequence to guarantee stable quality of reconstructed frames. 

\figurename~\ref{fig:inter_results_with_unified} provides R-D curves of the proposed method and state-of-the-art learning-based methods based on the unified transform framework under the PSNR metric. Following AlphaVC-cI~\cite{shi2022alphavc}, we encode one GOP of 96 frames for all the test sequences. The proposed method significantly outperforms existing unified-transform methods. This notable gain bridges the gap between the unified and separate transform frameworks, and highlights the superior efficiency of the proposed method. 
\figurename~\ref{fig:inter_results_psnr} and \figurename~\ref{fig:inter_results_ssim} compare the R-D curves with state-of-the-art learning-based separate-transform methods with the distortion measured by PSNR and MS-SSIM, respectively. Following DCVC-DC~\cite{Li2023-CVPR}, we encode 96 frames with a GOP size of 32 for all the test sequences.

\begin{figure}[!t]
\renewcommand{\baselinestretch}{1.0}
\setlength{\abovecaptionskip}{0pt}
\centering
\includegraphics[width=3.4 in]{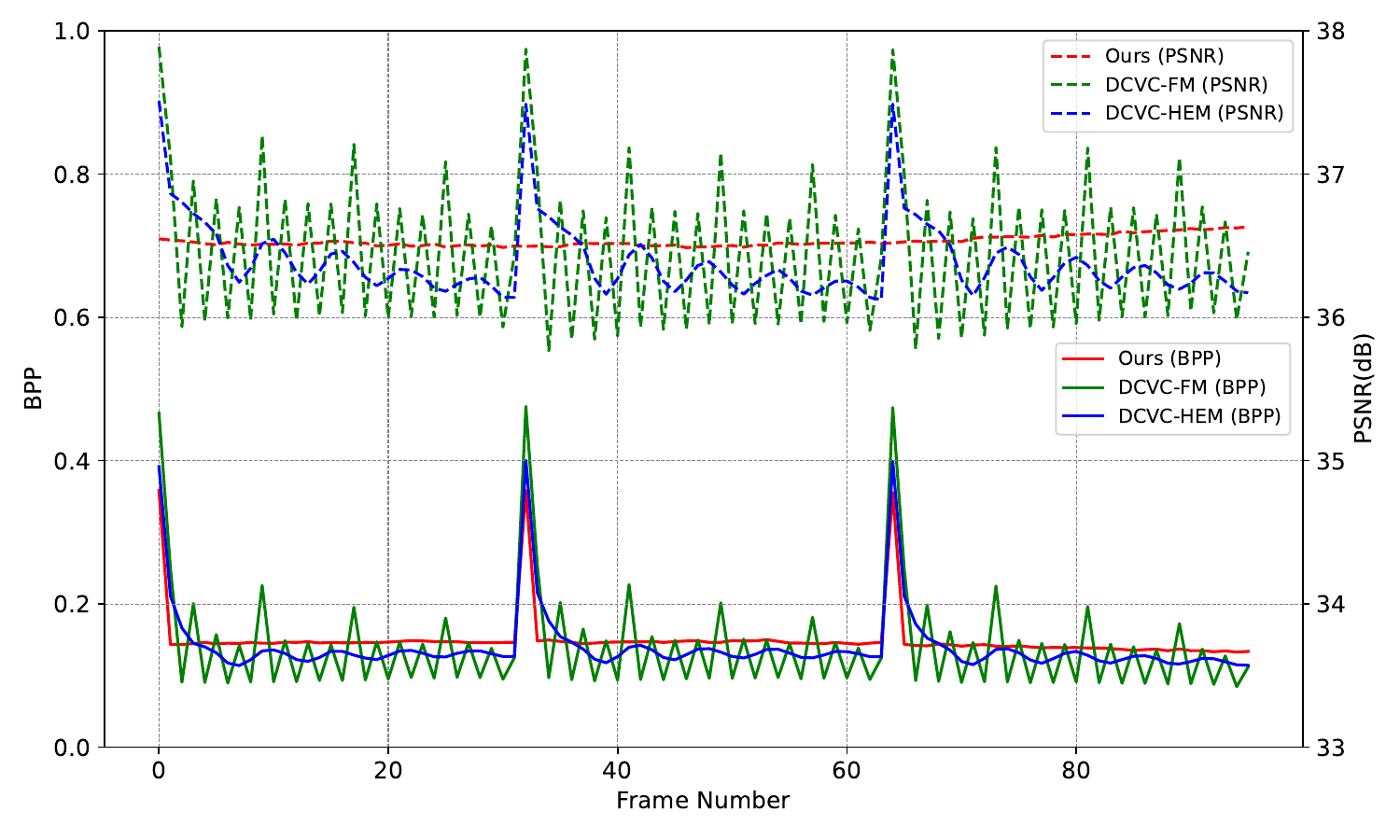}
\caption{Comparison on error propagation between the proposed method, DCVC-HEM~\cite{Li2021-NeurIPS} and DCVC-FM. We test 96 frames from the \emph{ShakeDry} sequence in the UVG~\cite{Mercat2020-UVG} dataset with the GOP size of 32. The results of DCVC-HEM and DCVC-FM are obtained by running their official code and pre-trained models. The proposed method is free from error propagation and maintains stable quality of decoded frames.}\label{fig:error_propagation}
\end{figure}

\begin{table}[!t]
\renewcommand{\baselinestretch}{1.0}   
\renewcommand{\arraystretch}{1.0}
\setlength{\abovecaptionskip}{0pt}
\centering
\caption{Comparison of computational and model complexity for compressing 1080p videos.}\label{tab:complexity}
\begin{threeparttable}
\begin{tabular}{@{}l cccc@{}}
\toprule
\multirow{2}{*}{Methods } &  Encoding & Decoding & \multirow{2}*{KMACs/pixel} & \multirow{2}*{\# Params (M)}\\ 
&(second)&(second)&&\\
\midrule
DCVC& 9.241 & 52.364 &1051& 36.2\\ 
DCVC-TCM&  0.773 &0.543&1462&46.5\\ 
DCVC-HEM&  0.643 &0.503&1621 &50.9\\
DCVC-DC&  0.794&0.612&1307 &50.8\\
DCVC-FM& 0.752 &0.584 &1103 &44.9 \\
VCT & 1.564 &1.421 & 2980&187.8 \\
Proposed & 0.729 & 0.605 &1426&  44.6\\ 
\bottomrule
\end{tabular}
\begin{tablenotes}
\item[$\ast$] The time of entropy coding is included. All the methods use ANS provided by CompressAI~\cite{Begaint2020-Arxiv} for entropy coding.
\end{tablenotes}
\end{threeparttable}
\vspace{3pt}
\renewcommand{\baselinestretch}{1.0}   
\renewcommand{\arraystretch}{1.0}
\setlength{\abovecaptionskip}{0pt}
\centering
\caption{Ablation study on latent scaling methods. KMACs per pixel and number of parameters are reported for intra-frame compression. BD-rates are calculated using naive latent scaling as the anchor.}\label{tab:ablation_variable}
\begin{tabular}{@{}l ccc@{}}
\toprule
Methods & BD-rates &KMACs/pixel &\# Params (M) \\ 
\midrule
Naive Latent Scaling~\cite{Chen2020-ICASSP} &0&405.7 &30.3\\
Channel-wise~\cite{Cui2021-CVPR} &-3.4\%&405.7&30.3\\
Content-Adaptive ~\cite{Li2022-ACMMM} &-8.6\%&415.3&33.4\\
Proposed QCMoE &-10.9\%&407.3&32.5\\
\bottomrule
\end{tabular}
\end{table}
\begin{figure*}[!t]
\renewcommand{\baselinestretch}{1.0}
\setlength{\abovecaptionskip}{0pt}
\centering
\includegraphics[width=7 in]{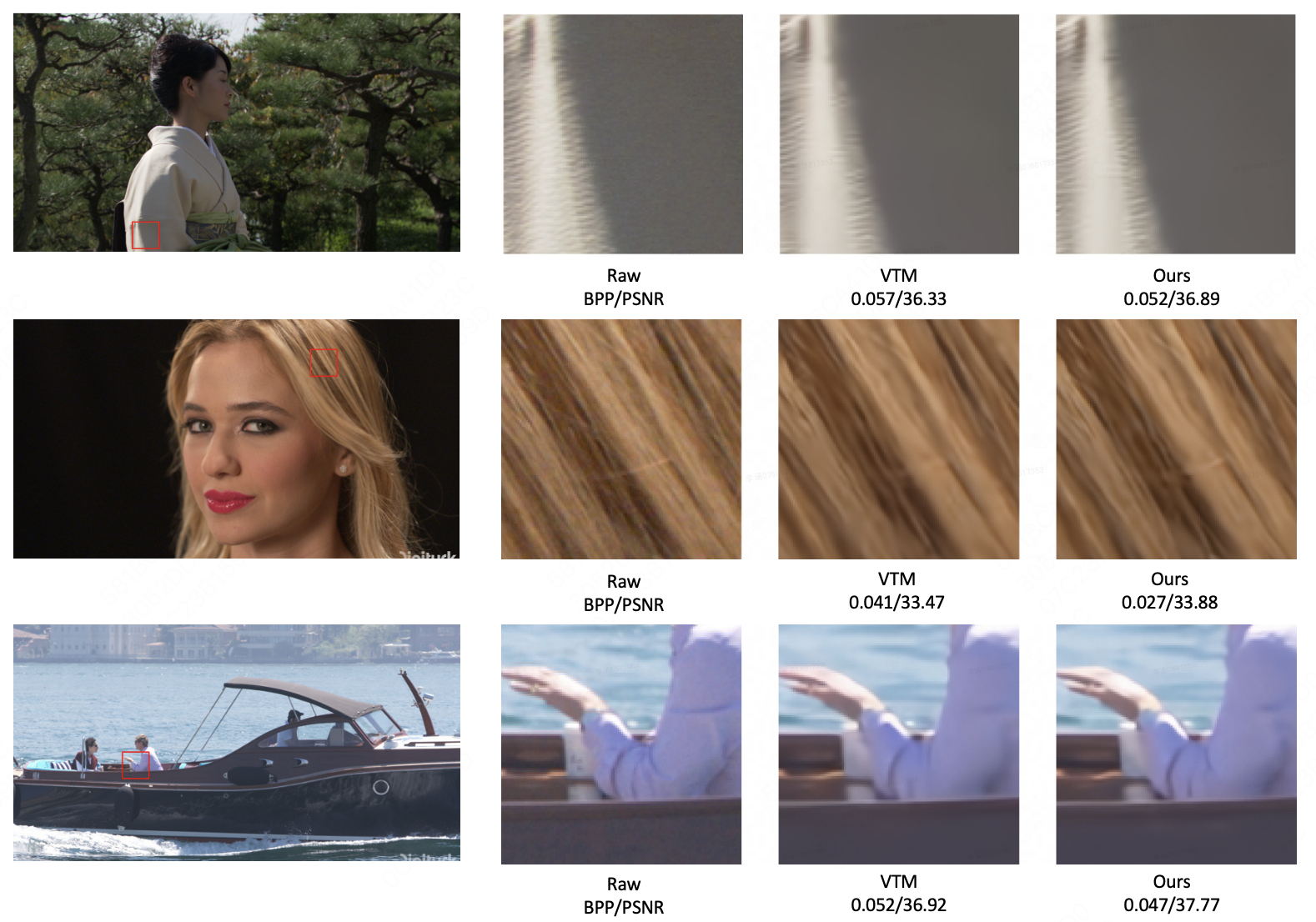}
\caption{The visual quality for the MSE-optimized models of the proposed method in comparison with VTM.}\label{fig:visual_quality}
\end{figure*}

Tables~\ref{tab:inter_results_psnr} and~\ref{tab:inter_results_psnr_ms-ssim} demonstrate that the proposed method achieves comparable R-D performance to state-of-the-art separate-transform methods in terms of BD-rates under the PSNR and MS-SSIM metrics. More importantly, \figurename~\ref{fig:error_propagation} shows that the proposed method has the significant advantage that it is \textbf{free from error propagation} compared with separate-transform methods. It achieves a BD-rate reduction of 10.4\% compared with VTM and outperforms DCVC-FM~\cite{li2024neural} by 3.2\% under the PSNR metric on MCL-JVC, the representative high-resolution test sequence with complex motion patterns, and is shown effective in handling challenging sequences with intricate motion dynamics. For low-resolution sequences with limited motion dynamics (\emph{e.g.}, HEVC Class C, D, E), the proposed method yields slightly lower R-D performance than DCVC-DC and DCVC-FM. 

\subsubsection{Error Propagation}
We compare the proposed method with the state-of-the-art methods (\emph{i.e.}, DCVC-HEM and DCVC-FM) and demonstrate that it can avoid error propagation. Experiments are performed on the first 96 frames from the \emph{ShakeDry} sequence in UVG~\cite{Mercat2020-UVG}. The results of DCVC-HEM and DCVC-FM are obtained by running their official implementation and pre-trained model. We keep almost the same average bit-rates by adjusting the quality scales of DCVC-HEM and DCVC-FM and the quality embedding of the proposed method. \figurename~\ref{fig:error_propagation} shows that DCVC-HEM suffers from serious error propagation, and is evidently degraded in reconstructing P-frames over time. DCVC-FM mitigates error propagation, but remains highly unstable in the quality of reconstructed P-frames and suffers from an evident gap of 2 dB compared to I-frames, which significantly affects the visual perception. In contrast, the proposed method is free from error propagation and maintains the stable reconstruction of each frame with comparable overall R-D performance to separate-transform methods.

\subsubsection{Visual Quality}
\figurename~\ref{fig:visual_quality} compares the visual quality of the proposed method optimized for MSE  with VTM. Our models achieve higher compression ratio and better visual quality than VTM.

\subsubsection{Model Complexity and Running Time}
Table~\ref{tab:complexity} reports the model complexity and running time for the proposed model and other state-of-the-art learned video compression methods. The proposed model has a total number of 46.9M parameters, including 32.5M for the intra-frame compression scheme and 14.4M for the inter-frame compression scheme (excluding shared transforms). The proposed model (1426 KMACs/pixel) is significantly lower in computational complexity than the unified-transform method VCT (2980 KMACs/pixel) and comparable to DCVC-DC (1307 KMACs/pixel) and DCVC-FM (1103 KMACs/pixel). The proposed method also demonstrates lower encoding and decoding time than VCT and DCVC, which is also comparable to DCVC-DC and DCVC-FM. This highlights the efficiency of our approach in balancing computational complexity and coding speed. The running time of the inter-frame compression scheme is evaluated using 1080p videos on a workstation equipped with a single NVIDIA 3090 GPU (24GB memory).

\subsection{Ablation Studies and Analysis}
\subsubsection{Ablation Study on QCMoE}
We perform ablation studies on our variable-rate compression methods to demonstrate the effect of the proposed QCMoE. Here, we only conduct the training stage I (\emph{i.e.}, intra-frame training), since the reconstruction quality is independent on the inter-frame compression.
All the experiments are evaluated on UVG and all the frames are regarded as I-frames.

\begin{table}[!t]
\renewcommand{\baselinestretch}{1.0}   
\renewcommand{\arraystretch}{1.0}
\setlength{\tabcolsep}{14pt}
\setlength{\abovecaptionskip}{0pt}
\centering
\caption{Ablation study on the number of experts. The MACs and model size only include the intra-frame compression part. The BD-rate is calculated using the anchor with $M=1$ and $K=1$.}\label{tab:ablation_qcmoe}
\begin{tabular}{@{}c c |ccc@{}}
\toprule
$M$ & $K$& BD-rates &KMACs/pixel &\# Params (M)\\ 
\midrule
1 & 1 & 0  & 406.5 & 30.6\\ 
6 & 1 & -3.7\%&406.5&32.5\\
6 & 2 & -6.2\% &407.3&32.5\\
6 & 4 & -6.9\%&408.9&32.5\\ 
4 & 2 &-5.3\% &407.3&31.7\\
6 & 2 &-6.2\% &407.3&32.5\\
12 & 2& -6.5\%& 407.3&34.9\\
\bottomrule
\end{tabular}
\vspace{3pt}
\setlength{\tabcolsep}{10pt}
\setlength{\abovecaptionskip}{0pt}
\centering
\caption{Ablation study on progressive alignment}\label{tab:abla-on-progressive-alignment}
\begin{tabular}{@{}cc|cc@{}}
\toprule
\multirow{2}{*}{Coarse Alignment} & \multirow{2}{*}{Refined Alignment} &  \multicolumn{2}{c}{BD-rates} \\ 
\cmidrule(ll){3-4} 
&& UVG & HEVC Class B \\ 
\midrule
 $\times$&$\times$ &0 &0 \\ 
\checkmark & $\times$&-11.5 & -7.7 \\ 
$\times$& \checkmark & -21.7 & -22.6\\ 
\checkmark  & \checkmark & -30.3& -31.3 \\
\bottomrule 
\end{tabular}
\vspace{3pt}
\setlength{\tabcolsep}{20pt}
\setlength{\abovecaptionskip}{0pt}
\centering
\caption{Ablation study on each component of proposed refined  alignment}\label{tab:abla-on-refiend-alignment}
\begin{tabular}{@{}cc|cc@{}}
\toprule
\multirow{2}{*}{LTMR} & \multirow{2}{*}{FG-DCA} &   \multicolumn{2}{c}{BD-rates} \\ 
\cmidrule(ll){3-4} 
&& UVG & HEVC Class B\\ 
\midrule
\checkmark &\checkmark &0 &0\\ 
$\times$ &\checkmark & +5.4 & +4.7\\ 
\checkmark &$\times$ & +11.3& +14.6\\
\bottomrule 
\end{tabular}
\end{table}

\textbf{Latent Scaling Methods.} 
We compare the proposed method with naive latent scaling~\cite{Chen2020-ICASSP}, channel-wise latent scaling~\cite{Cui2021-CVPR}, and content-adaptive latent scaling~\cite{Li2022-ACMMM}. Table~\ref{tab:ablation_variable} shows that, when deploying the proposed QCMoE (and \textit{i}-QCMoE), the proposed variable-rate intra-frame compression outperforms all the other latent scaling methods in R-D performance. Compared with the content-adaptive latent scaling of DCVC-HEM~\cite{Li2022-ACMMM}, the proposed QCMoE obtains better R-D performance with reduced computational and model complexity, demonstrating the superiority of dynamical allocation strategy of the proposed QCMoE module.

\textbf{Number of Experts.}
We further conduct experiments to diagnose the effect of the number of experts, including the total number of experts $M$ and the number of activated  experts  for each pixel $K$. Table~\ref{tab:ablation_qcmoe} shows that increasing both the total number of experts and the number of activated experts can enhance the model's ability to capture diverse patterns for variable-rate compression and enhance the overall R-D performance. To balance computational complexity and R-D performance, we choose $M=6$ and $K=2$ in this paper.

\subsubsection{Ablation Study on Inter-Frame Coding}
We further perform ablation studies to better understand how each component (including coarse alignment and refined alignment) of our inter-frame compression scheme affects the performance of video coding. 
For efficient comparison, we trained all variants for only 0.5 M steps in stage III. The BD-rate is evaluated on the UVG and HEVC Class B datasets and measured in PSNR.

\begin{figure}[!t]
\renewcommand{\baselinestretch}{1.0}
\setlength{\abovecaptionskip}{0pt}
\centering
\includegraphics[width=0.4\textwidth]{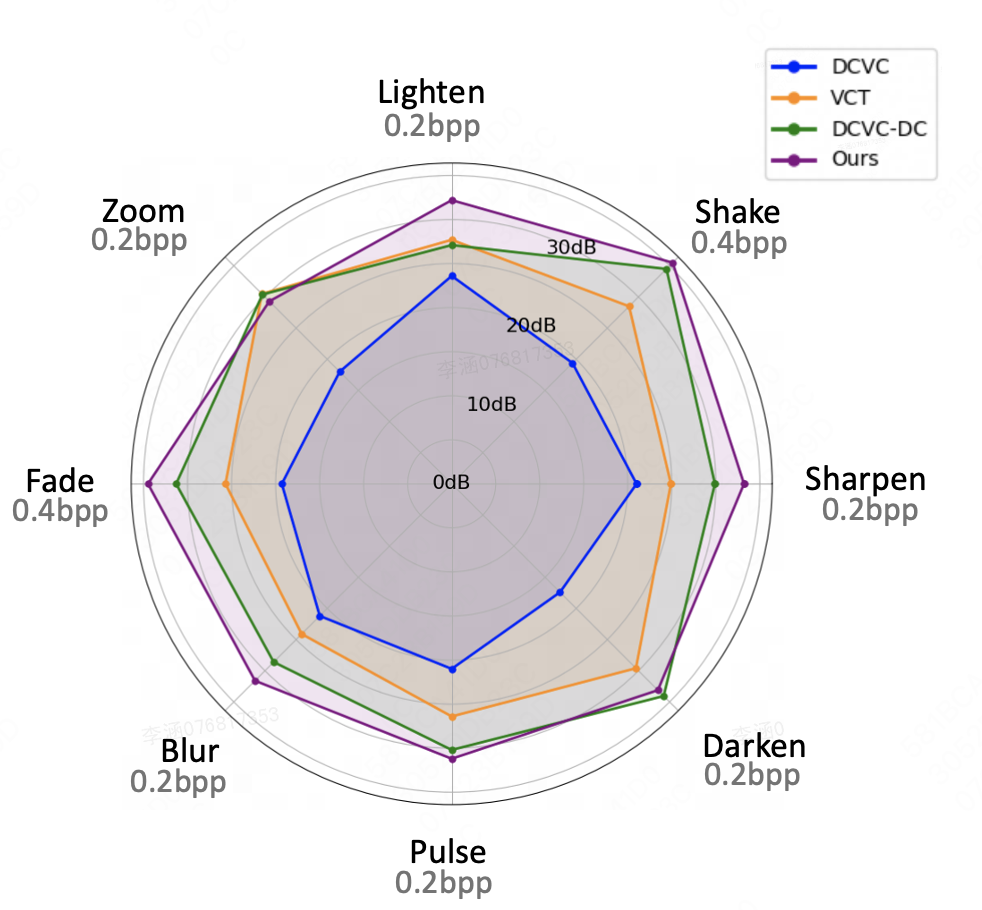}
\caption{The influence of diverse synthetic motion patterns on compression performance of, evaluated using PSNR values at an same bit-rate.}\label{fig:motion_generalization}
\renewcommand{\baselinestretch}{1.0}
\setlength{\abovecaptionskip}{0pt}
\centering
\includegraphics[width=2.5 in]{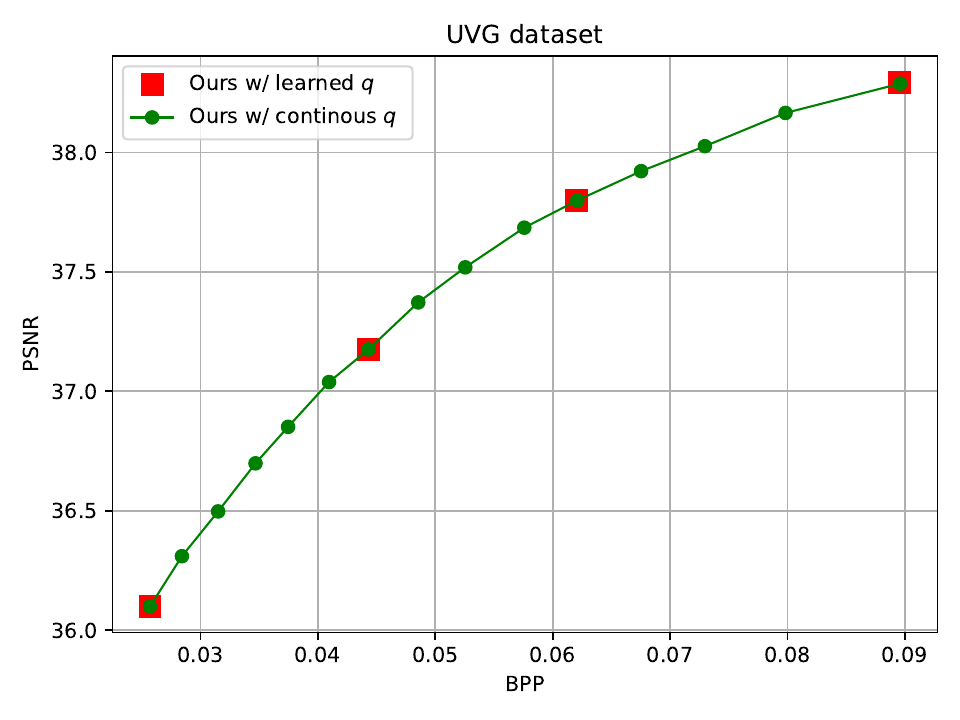}
\caption{Continuous rate adaptation using single model. We employ exponential interpolation based on the learned $\vect{q}$ to generate the continuous $\vect{q}$.}\label{fig:continous}
\end{figure}

\textbf{Proposed Progressive Alignment.} We conduct experiments with three variants of our proposed method to diagnose how the proposed progressive alignment method affects the R-D performance, including {\it i) No Alignment} that removes both the coarse pixel-domain alignment and refined latent-domain alignment and uses only the decoded latent of the last frame $\widehat{\vect{Y}}_{t-1}$ as the input to the inter-frame entropy model, serving as a temporal prior, {\it ii) Coarse Alignment Only} that removes the proposed refined latent-domain alignment and uses solely the coarse alignment latent as the input to the inter-frame entropy model, and {\it iii) Refined Alignment Only} that removes the proposed coarse pixel-domain alignment and regards the decoded latent of the last frame $\widehat{\vect{Y}}_{t-1}$ as the coarsely aligned latent, \emph{i.e.,} $\Tilde{\vect{Y}}_{t}=\widehat{\vect{Y}}_{t-1}$. Table~\ref{tab:abla-on-progressive-alignment} shows that removing both coarse alignment and refined alignment causes significant performance loss. Specifically, \emph{Coarse Alignment Only} achieves the BD-rate reduction of 11.5\% on UVG and 7.7\% on HEVC Class B compared to \emph{No Alignment}, indicating that coarse alignment alone provides limited improvement. \emph{Refined Alignment Only} improves BD-rate reduction to 21.7\% on UVG and 22.6\% on HEVC Class B, showing better performance than only using coarse alignment. \textit{Combined Progressive Alignment} significantly outperforms all the variants and achieves 30.3\% BD-rate reduction on UVG and 31.3\% on HEVC Class B. This highlights the complementary benefits of combining coarse alignment for capturing overall motion dynamics and refined alignment for optimizing detailed alignment accuracy.

\textbf{Each Component of Proposed Refined Alignment.}
We further investigate the contributions of each component of the proposed flow-guided refined latent-domain alignment to the final R-D performance. We conduct experiments using two variants of our method, including \textit{i) w/o LTMR} that removes the long-term motion refinement module and directly utilizes the downsampled optical flow  $\widehat{\vect{O}}_{t-1\rightarrow t}$ for the subsequent flow-guided deformable transformer and \textit{ii) w/o FG-DCA} that removes the optical flow guidance and employs standard cross-attention in~\eqref{eq:FG-DCA} instead of deformable cross-attention. Table~\ref{tab:abla-on-refiend-alignment} indicates that removing the long-term motion refinement (LTMR) leads to a BD-rate increase of 5.4\% and 4.7\% for UVG and HEVC Class B, respectively. Furthermore, removing the flow-guided deformable cross-attention (FG-DCA) results in more substantial BD-rate increase of 11.3\% for UVG and 14.6\% for HEVC Class B.
These results demonstrate that both long-term temporal information mining and flow-guided alignment significantly enhance the accuracy of aligned latent representations and further reduce temporal redundancy.

\subsubsection{Generalization to Various Motion Patterns}
Motion generalization is critical to handle different motion patterns in video coding. We follow~\cite{mentzer2022vct,lu2024high} to apply motion synthesis to the CLIC2020 test dataset~\cite{clic2021} and generate video sequences of 32 frames. For each sequence, we apply a specific type of motion to each frame based on its index $t=0,\cdots,31$.
\begin{itemize}
\item \textbf{Sharpening:} Progressively sharpening the $t$-th frame using a kernel size $(t/32)+1$ ranging from 1 to 2. 
\item \textbf{Lightening:} Gradually increasing the brightness of the $i$-th at a rate of \(t/32 \).
\item \textbf{Zooming:} Panning the $t$-th frame from left to right with a step size of \( 100\!\times\!(t/32) \) and scaling down by a factor of 0.98.
\item \textbf{Fading:} Fading the $t$-th frame at a rate of $t/32$.
\item \textbf{Blurring:} Applying progressive blurring effect to the $i$-th frame with the kernel size \( 2 \times \left\lfloor 5t/64 \right\rfloor + 1 \) increasing from 1 to 5 in steps of 2.
\item \textbf{Darkening:} Gradually reducing the brightness of the $t$-th frame at a rate of $1 - (t/32)$.
\item \textbf{Pulsing:} Applying brightness fluctuation with a period of 2 frames to the whole sequence.
\item \textbf{Shaking:} Simulated by random translations with a maximum amplitude of 20 pixels in any direction.
\end{itemize}

We compare the PSNR of reconstruction frames for various synthetic motion patterns under a fixed bit-rate. \figurename~\ref{fig:motion_generalization} shows that the proposed unified-transform framework with  progressive alignment outperforms other methods across most synthetic datasets, and is robust to handle diverse motion patterns in real-world scenarios for video compression.

\subsubsection{Continuous Rate Adaptation}
Table~\ref{tab:ablation_variable} shows the R-D performance gain by QCMoE. However, the learned quality embedding  set  $\mathcal{Q} =\{ \vect{q}_1,\cdots,\vect{q}_N\}$ corresponds to only $N$ bit-rate points ($N=4$ in our work). 
To achieve continuous rate adaptation, we follow~\cite{Cui2021-CVPR} to choose two quality embeddings $\{\vect{q}_{m},\vect{q}_{n}\}$ with $0\le m<n \le N-1$, and generate new quality embedding $\vect{q}_{r}= \vect{q}_{m}^l \vect{q}_{n} ^{1-l}$ by employing exponential interpolation with the interpolation coefficient $l\in \mathbb{R}$ to control the bit-rate. \figurename~\ref{fig:continous} demonstrates that continuous rate adaptation is achieved using a single codec by changing the value of continuous real $l$.

\section{Conclusions}\label{sec:conclusion}
In this paper, we presented a novel learned video compression method that effectively bridges the performance gap between unified-transform and separate-transform frameworks. Our key contribution is a dual-domain progressive temporal alignment method, which combines coarse pixel-domain motion estimation with refined latent-domain alignment using a Flow-Guided Deformable Transformer (FGDT), enabling more accurate motion compensation while preserving the error-free propagation advantage of the unified-transform framework.
Additionally, we introduced the Quality-Conditioned Mixture-of-Experts (QCMoE) module, which dynamically adapts quantization based on both content complexity and quality embedding, achieving flexible and efficient variable-rate compression. Extensive experiments demonstrate that our method matches the rate-distortion performance of state-of-the-art separate-transform approaches while maintaining the reconstruction stability and consistency of unified-transform coding.
Future work may explore more efficient transformer-based alignment and adaptive expert selection strategies to further improve compression efficiency. Our framework provides a promising direction for practical learned video compression systems that balance performance, flexibility, and robustness.

\ifCLASSOPTIONcaptionsoff
  \newpage
\fi

\end{document}